\documentclass[letterpaper]{article}
\usepackage{proceed2e}
\usepackage[margin=1in]{geometry}

\usepackage{microtype}
\usepackage{graphicx}
\usepackage{subfigure}
\usepackage{booktabs} 

\usepackage{fancyhdr,amsmath,amsthm,amssymb,bm} 
\usepackage[square,numbers]{natbib}
\usepackage{graphicx,caption}
\usepackage{float}
\usepackage{indentfirst}
\usepackage{bm}
\usepackage{bbm}
\usepackage{amsfonts,verbatim}
\usepackage{dsfont}
\usepackage{color}
\usepackage{diagbox}
\usepackage[ruled]{algorithm2e}

\usepackage{hyperref}

\newcommand{\overbar}[1]{\mkern 1.5mu\overline{\mkern-1.5mu#1\mkern-1.5mu}\mkern 1.5mu}

\let\tilde\widetilde
\let\hat\widehat

\ifx\theorem\undefined
\newtheorem{theorem}{Theorem}
\fi
\ifx\proposition\undefined
\newtheorem{proposition}[theorem]{Proposition}
\fi
\ifx\proof\undefined
\newenvironment{proof}{\par\noindent{\bf Proof\ }}{\hfill\BlackBox\\[2mm]}
\fi

\graphicspath{{figs/}}
\graphicspath{{figs_uni/}}

\usepackage{times}

\title{Simultaneous Parameter Learning and Bi-Clustering\\ for Multi-Response Models}

\author{} 

\author{ {\bf Ming Yu} \\
Booth School of Business\\
The University of Chicago\\
Chicago, IL \\
ming93@uchicago.edu\\
\And
{\bf Karthikeyan Natesan}\\
\bf{Ramamurthy}\\
IBM Research\\
Yorktown Heights,  NY\\
knatesa@us.ibm.com\\
\And
{\bf Addie Thompson}\\
Dept. of Plant, Soil \\
and Microbial Sciences \\
Michigan State University\\
East Lansing,  MI\\
thom1718@msu.edu\\
\And
{\bf Aur\'{e}lie Lozano}  \\
IBM Research\\
Yorktown Heights,  NY\\
aclozano@us.ibm.com\\
}
\begin{document}

\maketitle

\begin{abstract}
We consider multi-response and multitask regression models, where the parameter matrix to be estimated is expected to have an \emph{unknown} grouping structure. The groupings can be along tasks, or features, or  both, the last one indicating a bi-cluster or ``checkerboard'' structure. Discovering this grouping structure along with parameter inference makes sense in several applications, such as  multi-response Genome-Wide Association Studies. This additional structure can not only can be leveraged for more accurate parameter estimation, but it also provides valuable information on the underlying data mechanisms (e.g. relationships among genotypes and phenotypes in GWAS). In this paper, we propose two formulations to simultaneously learn the parameter matrix and its group structures, based on convex regularization penalties. We present optimization approaches to solve the resulting problems and provide numerical convergence guarantees. Our approaches are validated on extensive simulations and real datasets concerning phenotypes and genotypes of plant varieties.
\end{abstract}

\section{INTRODUCTION} \label{sec:introduction}
\begin{figure}
\begin{center}
\centerline{\includegraphics[width=\columnwidth]{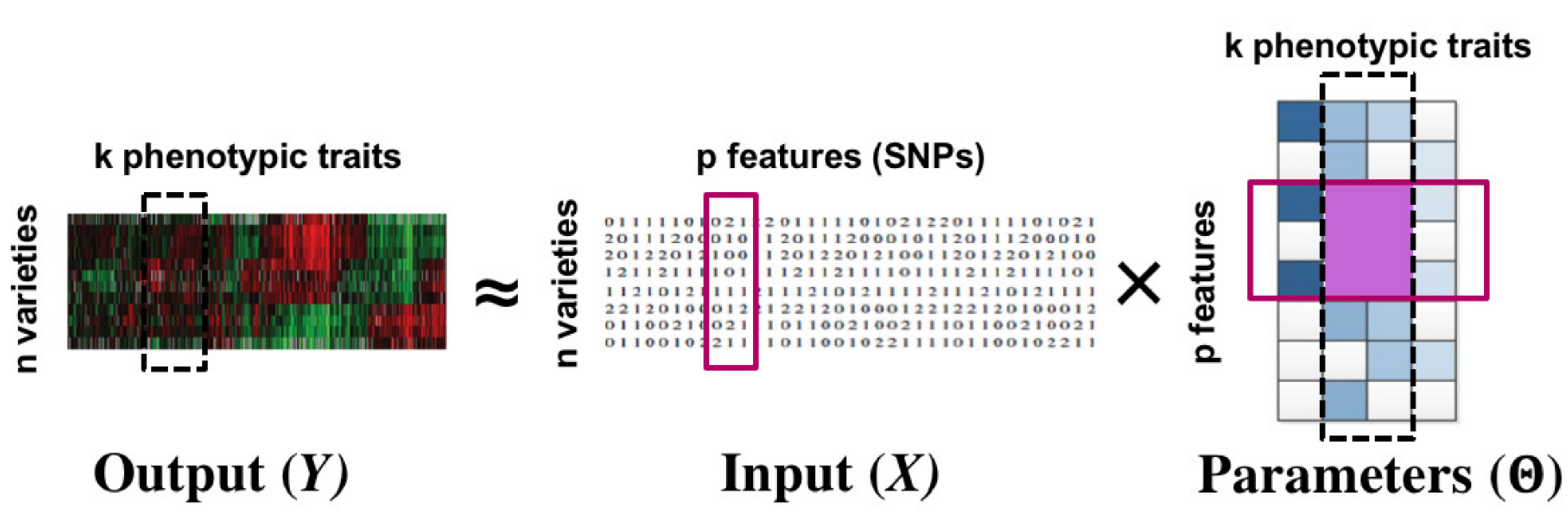}}
\caption{Multi-response GWAS: The simultaneous grouping relationship between phenotypic traits and SNPs manifest as a block structure (row $+$ column groups) in the parameter matrix. The row and column groups are special cases of the more general block structure. Our proposed approach infers the parameter matrix as well as the group structures.}
\label{fig:multi_response_GWAS}
\end{center}
\end{figure}

We consider multi-response and multi-task regression models, which generalize single-response regression to learn predictive relationships between multiple input and multiple output variables, also referred to as tasks \cite{borchani2015survey}. The parameters to be estimated form a matrix instead of a vector. In many applications, there exist grouping structures among input variables and output variables, and the model parameters belonging to the same input-output group tend to be close to each other.  A motivating example is that of multi-response Genome-Wide Association Studies (GWAS)~\cite{schifano2013genome}, where for instance a group of Single Nucleotide Polymorphisms or SNPs (input variables or features) might influence a group of phenotypes (output variables or tasks) in a similar way, while having little or no effect on another group of phenotypes. It is therefore desirable to \emph{uncover and exploit} such input-output structures in estimating the parameter matrix. See figure \ref{fig:multi_response_GWAS} for an example.

\paragraph{Contributions:} In this work, we develop formulations that \emph{simultaneously} learn: (a) the parameters of multi-response/task regression models, and, (b) the grouping structure in the parameter matrix (row or column or both) that reflects the group relationship between input and output variables. We present optimization approaches to efficiently solve the resulting convex problems, and show their numerical convergence. We describe and justify various options we take and choices we make in the optimization algorithms. Our proposed methods are validated empirically on synthetic data and on a real-world datasets concerning phenotypes and genotypes of plant varieties. From the synthetic data experiments, we find that our methods provide a much better and more stable (i.e., lesser standard error) recovery of the underlying group structure, and improved estimates of parameters. In real-world data experiments, our approaches reveal natural groupings of phenotypes and \textit{checkerboard} patterns of phenotype-SNP groups that inform us of the joint relationship between them. In all experiments, we demonstrate better RMSEs.

We emphasize that the parameters as well as the grouping structures are \textit{fully unknown} a-priori, and inferring them simultaneously is our major contribution. This is in contrast to the naive way of estimating the parameters first and then clustering. This naive approach has the danger of propagating the estimation error into clustering, particularly in high dimensions, where the estimator is usually inaccurate due to lack of sufficient samples. Moreover, the clustering step of the naive approach does not use the full information of the data. The joint estimation-clustering procedure we propose naturally promotes sharing of information within groups. Our formulations adopt the convex bi-clustering cost function \cite{chi2014convex} as the regularizer to encourage groupings between columns (tasks) and rows (features) in the parameter matrix. Note that \cite{chi2014convex} assume that the data matrix to be used for bi-clustering is known a-priori, which is obviously not the case for our setting. 


\paragraph{Related Work:} Multi-task learning attempts to learn several of the inference tasks simultaneously, and the assumption here is that an appropriate sharing of information  can benefit all the tasks \cite{caruana1998multitask, obozinski2006multi, yu2018recovery}. 
The implicit assumption that all tasks are closely related can be excessive as it ignores the underlying specificity of the mappings. There have been several extensions to multi-task learning that address this problem. The authors in \cite{jalali2010dirty} propose a \emph{dirty} model for feature sharing among tasks, wherein a linear superposition of two sets of parameters - one that is common to all tasks, and one that is task-specific - are used. \cite{kim2010tree} leverages a \emph{predefined} tree structure among the output tasks (e.g. using hierarchical agglomerative clustering) and imposes group regularizations on the task parameters based on this tree. The approach proposed in \cite{kumar2012learning} learns to share by defining a set of \emph{basis task parameters} and posing the task-specific parameters as a sparse linear combination of these. The approaches of  \cite{jacob2009clustered} and \cite{kang2011learning} assume that the tasks are clustered into groups and proceed to learn the group structure along with the task parameters using a convex and an integer quadratic program respectively. However, these approaches do not consider joint clustering of the features. In addition, the mixed integer program of \cite{kang2011learning} is computationally intensive and greatly limits the maximum number of tasks that can be considered. Another pertinent approach is the Network Lasso formulation presented in~\cite{hallac2015network}. This formulation, however, is limited to settings where only clustering among the tasks is needed. A straightforward special case of the proposed approach to column- or row-only clustering (a.k.a. \emph{Uni-clustering}) is presented in \cite{Yu2017multitask}. Portions of \cite{Yu2017multitask} are reproduced in this paper also for comprehensive coverage.


\paragraph{Roadmap.} In Section \ref{sec:formulation}, we will discuss the proposed joint estimation-clustering formulations, and in Section \ref{sec:optimization}, we will present the optimization approaches. The choice of hyperparameters used and their significance is discussed in Section \ref{sec:procedure}. We illustrate the solution path for one of the formulations in Section \ref{sec:solution_path}. We will provide results for estimation with synthetic data, and two case studies using multi-response GWAS with real data in Sections \ref{sec:simulation} and \ref{sec:real_data} respectively. We conclude in Section \ref{sec:conclusion}. Additional details, convergence proofs, solution paths, and experiments are provided in the supplementary material.




\section{PROPOSED FORMULATIONS} \label{sec:formulation}
We will motivate and propose two distinct formulations for simultaneous parameter learning and clustering with general supervised models involving matrix valued parameters. Our formulations will be developed around multi-task regression in this paper.



We let $X_s \in \mathbb{R}^{n\times p}$ be the design matrices and $Y_s \in \mathbb{R}^n$ be the response vectors for each task $s=\{1,\ldots,k\}$, in multi-task regression. $\Theta \in \mathbb{R}^{p\times k}$ is the parameter or coefficient matrix for the $k$ tasks. We wish to simultaneously estimate $\Theta$ and discover the bi-cluster structure among features and tasks, respectively the rows and columns of $\Theta$. Note that discovering groupings just along rows or columns is a special case of this.

%
%

\subsection{Formulation 1:}
\label{sec:form1}
We begin with the simplest formulation, which, as we shall see, is a special case of the latter one.
\begin{equation}
\small
\begin{aligned}
\min_{\Theta} \, L(X, Y; \Theta) + \lambda_1 R(\Theta) + \lambda_2 \Big[ \Omega_W(\Theta) + \Omega_{\tilde W}(\Theta^T) \Big].
\end{aligned}
\label{eqn:form_1_general}
\end{equation} Here $L(X, Y; \Theta)$ is the loss function, $R(\Theta)$ is a regularizer, and $\Omega_W(\Theta) = \sum_{i<j} w_{ij}\| \Theta_{\cdot i} - \Theta_{\cdot j} \|_2$ and $ \Theta_{\cdot i} $ is the $i^{\text{th}}$ column of $ \Theta $.  $\Omega_W(\Theta)$ is inspired by the convex bi-clustering objective \cite[eqn. 2.1]{chi2014convex} and it encourages sparsity in differences between columns of $\Theta$. Similarly,  $\Omega_W(\Theta^T)$ encourages sparsity in the differences between the rows of $\Theta$. When the overall objective is optimized, we can expect to see a checkerboard pattern in the model parameter matrix. Note that $W$ and $\tilde W$ are nonnegative weights that reflect our prior belief on the closeness of the rows and columns of $\Theta$.

\begin{figure}
\centering
\begin{minipage}[t]{0.3\linewidth}
\centering
\includegraphics[width=\textwidth]{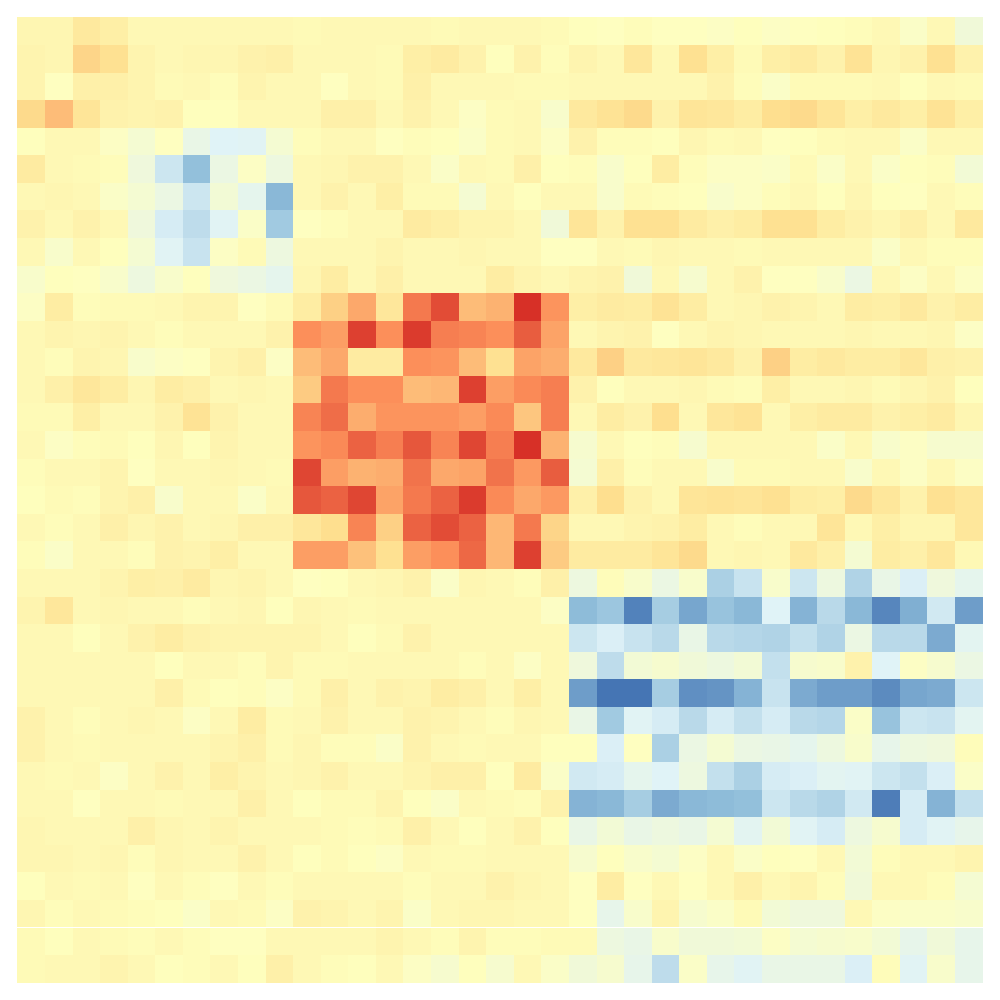}
\end{minipage}
\begin{minipage}[t]{0.3\linewidth}
\centering
\includegraphics[width=\textwidth]{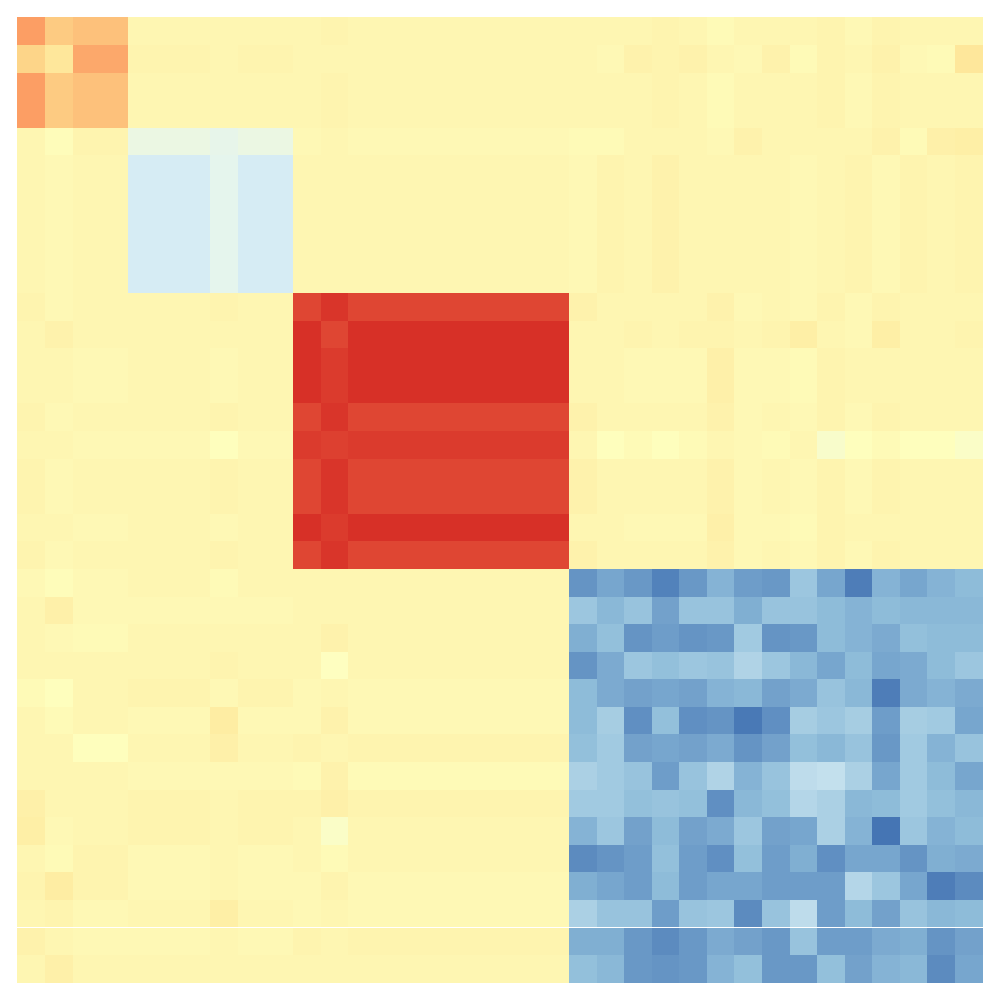}
\end{minipage}

\begin{minipage}[t]{0.3\linewidth}
\centering
\includegraphics[width=\textwidth]{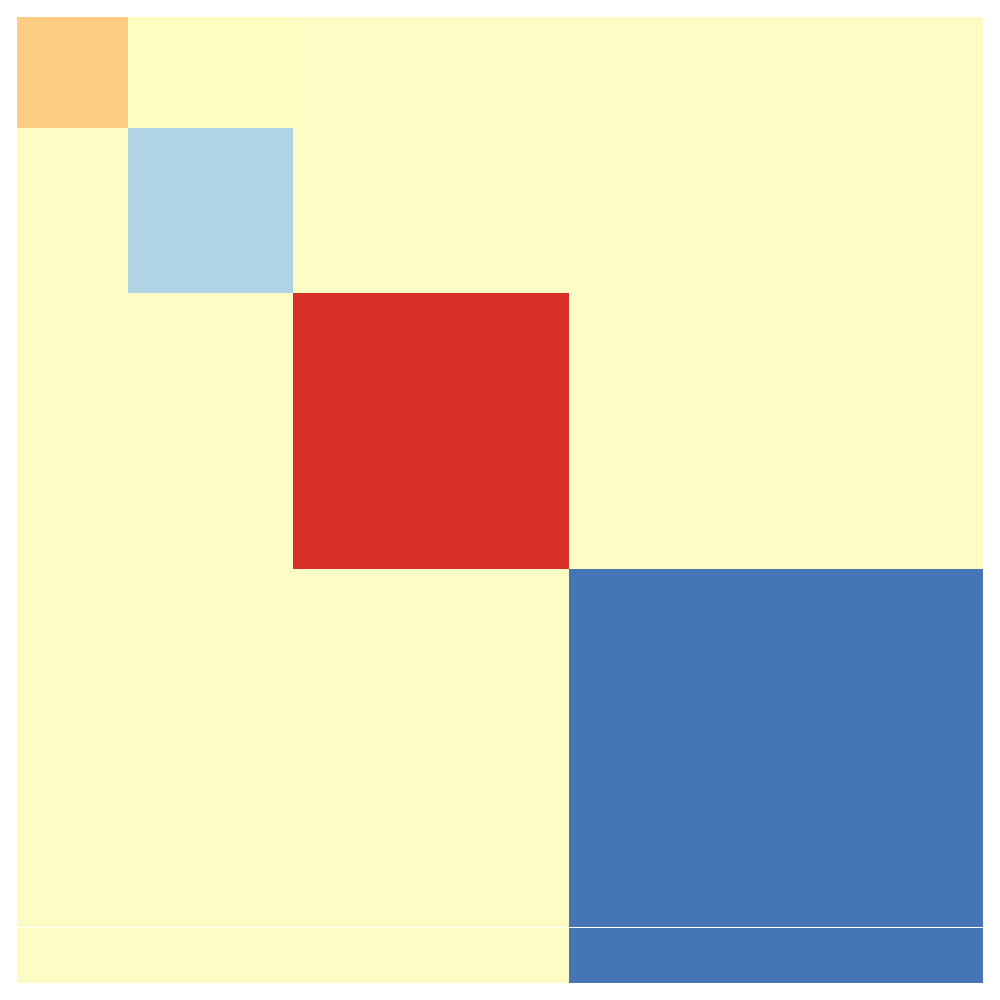}
\end{minipage}
\begin{minipage}[t]{0.3\linewidth}
\centering
\includegraphics[width=\textwidth]{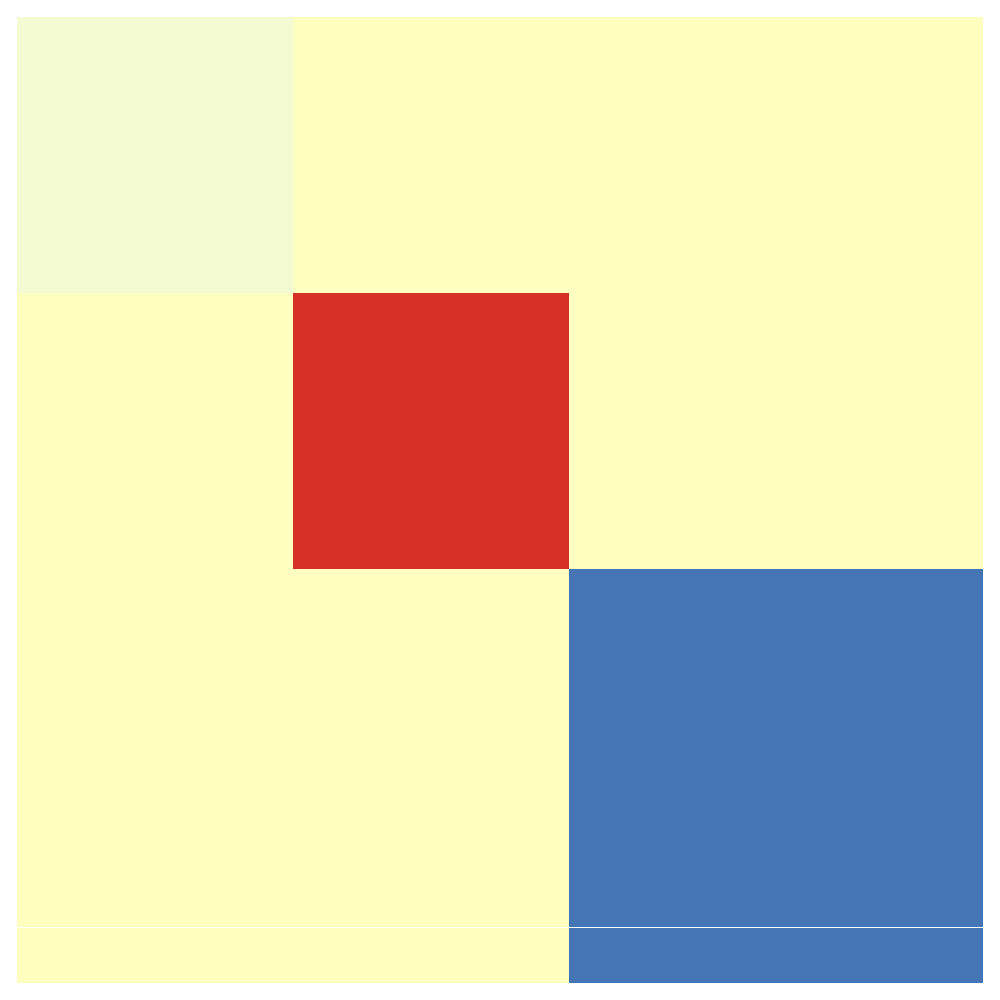}
\end{minipage}
\caption{Evolution of bi-clustering structure of $\Theta$ as $\lambda_2$ increases in the order top left, top right, bottom left, bottom right.}
\label{illustration}
\end{figure}

The degree of \emph{sharing} of parameters and hence that of bi-clustering, is controlled using the tuning parameter $\lambda_2$. When $\lambda_2$ is small, each element of $\Theta$ will be its own bi-cluster. As $\lambda_2$ increases, more elements of $\Theta$  \emph{fuse} together, the number of rectangles in the checkerboard pattern will reduce. See Figure \ref{illustration} for the change of the checkerboard structure as $\lambda_2$ increases. Further, by varying $\lambda_2$ we get a solution path instead of just a point estimate of $\Theta$. This will be discussed more in Section \ref{sec:solution_path}. In the remainder of the paper, we will use the same design matrix $X$ across all tasks for simplicity, without loss of generality.

For sparse multi-task linear regression, formulation 1 can be instantiated as,
\begin{equation}
\small
\begin{aligned}
&\min_{\Theta} \, \| Y - X \Theta\|_F^2  +  {\lambda_1  \sum_{i=1}^k \| \Theta_{\cdot i}\|_1} + \lambda_2 \Big[ \Omega_W(\Theta) + \Omega_{\tilde W}(\Theta^T) \Big].
\end{aligned}
\label{eqn:form_1_MTL}
\end{equation} Here the rows of $\Theta$ correspond to the features, i.e. the columns of $X$, and the columns of  $\Theta$ correspond to the tasks, i.e., the columns of $Y$. Therefore, the checkerboard pattern in $\Theta$ provides us insights on the groups of features that go together with the groups of tasks.

\subsection{Formulation 2}
\label{sec:form2}
Formulation 1 is natural and simple, but it forces the parameters belonging to the same row or column cluster to be equal, and this may be limiting. To relax this requirement, we introduce a surrogate parameter matrix $\Gamma$ that will be used for bi-clustering. This will be mandated to be close to $\Theta$. This yields the following objective
\begin{equation}
\begin{aligned}
&\min_{\Theta,\Gamma} \, \| Y - X \Theta\|_F^2  + \lambda_1  \sum_{i=1}^k \| \Theta_i\|_1  + \lambda_2 \sum_{i=1}^k \| \Theta_i - \Gamma_i\|_2^2 \\
&   \qquad   +  \lambda_3  \left[ \Omega_W(\Gamma) + \Omega_{\tilde W}(\Gamma^T) \right].
\end{aligned}
\label{eqn:form_2_MTL}
\end{equation} 

\paragraph{Remark 1.}
To interpret this more carefully, let us assume that  $\Theta_i = \overbar{\Theta_i} + \Gamma_i$ in (\ref{eqn:form_2_MTL}). In other words, $\Theta_i$ has a global component $\overline{\Theta_i}$, and the component $\Gamma_i$ that participates in the clustering. As $\lambda_2 \rightarrow \infty$, $\overbar{\Theta_i}  \rightarrow 0$, and hence $\Theta_i \rightarrow\Gamma_i$. Now, formulation 2 reduces to formulation 1. Further, if $\lambda_1$ and $\lambda_2$ are held constant while only $\lambda_3$ increases, $\Theta_i \rightarrow \overbar{\Theta_i} + \Gamma$, since $\Gamma_i \rightarrow \Gamma$ for all $i$. The key difference between formulation 2 and 1 is the presence of a task-specific global component $\overbar{\Theta_i}$, which lends additional flexibility in modeling the individual tasks even when $\lambda_3 \rightarrow 0$. Whereas, in (\ref{eqn:form_1_MTL}), when $\lambda_2 \rightarrow 0$, $\Theta_i \rightarrow \Theta$ for all $i$, and the tasks are forced to share the same coefficients without any flexibility.

\paragraph{Remark 2.}  In certain applications, it might make sense to cluster together features / tasks whose effects have the same amplitude but different signs. This can be accommodated by considering $\Omega_W(\Theta)  = \sum_{i<j} w_{ij}\| \Theta_{\cdot i} - c_{ij} \Theta_{\cdot j} \|_2$ where $c_{i,j} \in \{-1, 1\}$ are predefined constants reflecting whether the features or tasks are expected to be negatively or positively correlated.

\section{OPTIMIZATION APPROACHES} \label{sec:optimization}
We describe the optimization procedures to solve the two proposed formulations. Note that as long as the loss function $L(X, Y; \Theta)$ and the regularization $R(\Theta)$ are convex, our formulations are also convex in $\Theta$ and $\Gamma$, and hence can be solved using modern convex optimization approaches. Here we adopt two computationally efficient approaches.

\subsection{Formulation 1} \label{subsec:opt_f1}
For our formulation 1 we use the proximal decomposition method introduced in \cite{combettes2008proximal}. This is an efficient algorithm for minimizing the sum of several convex functions. Our general objective function \eqref{eqn:form_1_general} involves $3$ such functions: $f_1$ being $L(X,Y; \Theta)$, $f_2$ being $R(\Theta)$ and $f_3$ being the term that multiplies $\lambda_2$. At a high level, the algorithm iteratively applies proximal updates with respect to these functions until convergence. 

We stack the regression matrix $\Theta$ into a column vector $(\Theta_1; \ldots ; \Theta_k ) \in \mathbb R^{pk}$. The proximal operator is given by:
\begin{equation}
\text{prox}_f  b  = \mathop{\text{argmin}}\limits_a \Big( f(a) + \frac 12 \|b-a\|_2^2 \Big),
\label{proximal}
\end{equation}
where $a$ and $b$ are $pk$-dimensional vectors.
The proximal operator of the regularized loss can be computed according to the specific $L$ and $R$ functions. The overall optimization procedure is given in Algorithm \ref{alg:f1} and update rules are given in the supplementary material.



\begin{algorithm}
\SetAlgoLined
\DontPrintSemicolon
\KwResult{Estimated $\Theta$}
 Initialize $\tilde \Theta_{1,0}, \tilde \Theta_{2,0}, \tilde \Theta_{3,0} \in \mathbb{R}^{pk}$, $\gamma \in (0, \infty)$, $m = 0$ \;
 Calculate $\hat \Theta_0 = \frac 13(\tilde \Theta_{1,0} + \tilde \Theta_{2,0} + \tilde \Theta_{3,0})$ \;
 \While{not converged}{
   \For{$i = 1, 2, 3$}{
   $p_{i,m} = \text{prox}_{\gamma f_i}(\tilde \Theta_{i, m})$\;
   }
   $p_m = \frac 13(p_{1,m}+p_{2,m}+p_{3,m})$ \;
   \For{$i = 1, 2, 3$}{
   $\tilde \Theta_{i,m+1} = \tilde \Theta_{i,m} + 2p_m - \hat \Theta_m - p_{i,m}$ \;
   }
   $\hat \Theta_{m+1} = p_m$ \;
   $m = m+1$\;
 }
 Reshape $\hat \Theta_m$ to get estimated $\Theta$.
 \caption{Proximal decomposition for formulation 1}
 \label{alg:f1}
\end{algorithm}

\subsection{Formulation 2} \label{subsec:opt_f2}

%


For our formulation 2 we use an alternating minimization method on $\Theta$ and $\Gamma$; i.e., we alternatively minimize over $\Theta$ and $\Gamma$ with the other fixed. 

The first step in the alternating minimization is to estimate $\Theta$ while fixing $\Gamma$.
This minimization problem is separable for each column and each sub-problem can be easily written as a standard Lasso problem:
\begin{equation}
\min_{\Theta_i} \, \| \tilde y_i - \tilde X \Theta_i\|_2^2  + \lambda_1 \| \Theta_i\|_1  
\label{lasso_rewritten}
\end{equation}
by defining 
\begin{equation}
\tilde y_i = 
\begin{pmatrix}
y_i \\
\quad \\
\sqrt{\lambda_2} \Gamma_i
\end{pmatrix}
\quad \text{and} \quad
\tilde X = \begin{pmatrix}
X \\
\quad \\
\sqrt{\lambda_2} I_p
\end{pmatrix}
\label{lasso_define}
\end{equation}
and hence can be solved efficiently and in parallel for each column.

In the second step, we fix $\Theta$ and optimize for $\Gamma$. The optimization is
\begin{equation}
\label{eqn:optimize_form_2_MTL_Gamma}
\textrm{minimize}_{\Gamma}  \sum_{i=1}^k \| \Theta_i - \Gamma_i\|_2^2 +  \frac{\lambda_3}{\lambda_2}  \left[ \Omega_W(\Gamma) + \Omega_{\tilde W}(\Gamma^T) \right]
\end{equation} which is a standard bi-clustering problem on $\Theta$ and can be solved efficiently using the COnvex BiclusteRing  Algorithm (COBRA) introduced in \cite{chi2014convex}, and described in Algorithm \ref{alg:COBRA} in supplementary material for completeness. The overall procedure is given in Algorithm \ref{alg:f2}. 

\begin{algorithm}
\SetAlgoLined
\DontPrintSemicolon
\KwResult{Estimated $\Theta$ and $\Gamma$}
 Initialize $\Theta_0$, $\Gamma_0$, iteration $m = 0$ \;
 \While{not converged}{
   Estimate $\Theta_m$ by solving \eqref{lasso_rewritten} using LASSO \;
   Estimate $\Gamma_m$ by solving \eqref{eqn:optimize_form_2_MTL_Gamma} using COBRA \;
   $m = m+1$\;
 }
 \caption{Alternating minimization for formulation 2}
 \label{alg:f2}
\end{algorithm}
 
 
 

\subsection{Convergence}

We establish the following convergence result for our algorithms, when the loss function $L(X. Y; \Theta)$ is convex in $\Theta$. The proofs are given in the supplementary material.

\begin{proposition}
\label{prop:f1}
The algorithm described in Section \ref{subsec:opt_f1} converges to the global minimizer.
\end{proposition}

\begin{proposition}
\label{prop:f2}
The algorithm described in Section \ref{subsec:opt_f2} converges to the global minimizer.
\end{proposition}

\section{HYPERPARAMETER CHOICES AND VARIATIONS} 
\label{sec:procedure}
We will describe and justify the various choices for hyperparameters while optimizing formulations 1 and 2.

\subsection{Weights and Sparsity Regularization}
\label{sec:weights_sparsity}
The choice of the column and row similarity weights $W$ and $\tilde{W}$ can dramatically affect the quality of the clustering results and we follow the suggestion in \cite{chi2014convex} to set these. However, we need an estimate of $\Theta$ to obtain the weights and this can be found by solving 
\begin{equation}
\begin{aligned}
\min_{\Theta} \, \| Y - X \Theta\|_F^2  + \lambda_1  \sum_{i=1}^k \| \Theta_i\|_1,
\end{aligned}
\label{eqn:Lasso}
\end{equation} 
where $\lambda_1$ is tuned using cross-validation (CV) and re-used in the rest of the algorithm. From our multi-task regression experiments, we find that the clustering results are quite robust to the choice of $\lambda_1$.


With the estimated $\hat{\Theta}$ we use the approach suggested in \cite{chi2014convex} to compute $W$ and $\tilde{W}$. The weights for the columns $i$ and $j$ are computed as $w_{ij} = 1^k_{ij} \cdot \exp\big(-\phi\| \hat \Theta_{\cdot i} - \hat\Theta_{\cdot j} \|_2^2\big)$ where $1^k_{ij}$ is 1 if $j$ is among $i$'s $\kappa$-nearest-neighbors or vice versa and $0$ otherwise. $\phi$ is nonnegative and $\phi = 0$ corresponds to uniform weights. In our synthetic and real data experiments we fix $\phi = 20$. $\tilde{W}$ is computed analogously. It is important to keep the two penalty terms $\Omega_W(\Theta)$ and $\Omega_{\tilde W}(\Theta^T)$ on the same scale, else the row or column objective will dominate the solution. We require that the column weights sum to $1/\sqrt n$ and the row weights sum to $1/\sqrt p$, following \cite{chi2014convex}. More rationale on the weight choices is provided in \cite{chi2014convex} and \cite{chi2015splitting}.


\subsection{Penalty Multiplier Tuning}
\label{sec:penalty_multiplier_tuning}
We set the penalty multipliers ($\lambda_1$, $\lambda_2$, and $\lambda_3$) for both the formulations using a CV approach. 
We randomly split our samples into a training set and a hold-out validation set, fitting the models on the training set and then evaluating the root-mean-squared error (RMSE) on the validation set to choose the best values. In order to reduce the computational complexity, we estimate the multipliers greedily, one or two at a time. From our simulations, we determined that this is a reasonable choice. We recognize that these can be tuned further on a case-by-case basis with additional computational complexity.

$\lambda_1$ is set to the reasonable value as determined in Section \ref{sec:weights_sparsity} for both formulations, since the clustering results are quite robust to this choice. For formulation 1, we estimate the best $\lambda_2$ by CV using \eqref{eqn:form_1_general}. 
For formulation 2, the tuning process is similar, however we pick a sequence of $\lambda_2$ and $\lambda_3$. We estimate both $\hat \Theta_{\lambda_2, \lambda_3}$ and $\hat \Gamma_{\lambda_2, \lambda_3}$, but calculate RMSE with $\hat \Gamma_{\lambda_2, \lambda_3}$, since it directly participates in the clustering objective. When the path of bi-clusterings is computed, we fix $\lambda_2$ to the CV estimate and vary only $\lambda_3$.


\subsection{Bi-clustering Thresholds}
\label{sec:biclus_thresh}
It is well known that LASSO tends to select too many variables {\cite{meinshausen2009lasso}. Hence $\|\Theta_{\cdot i} - \Theta_{\cdot j}\|_2$ may not be exactly zero in most cases, and we may end up identifying too many clusters as well. In \cite{chi2014convex} the authors defined the measure $v_{ij} = \|\Theta_{\cdot i} - \Theta_{\cdot j}\|_2$ and placed the $i^{th}$ and $j^{th}$ columns in the same group if $ v_{ij} \leq \tau$ for some threshold $\tau$, inspired by \cite{meinshausen2009lasso}. In our formulation 1, after selecting the best tuning parameters and estimating $\Theta$, we place the $i^{th}$ and $j^{th}$ rows in the same group if $\| \Theta_{i \cdot} - \Theta_{j \cdot} \|_2 \leq \tau_r$. Similarly, if $\| \Theta_{\cdot i} - \Theta_{\cdot j} \|_2 \leq \tau_c$ we place the $i^{th}$ and $j^{th}$ columns in the same group. For formulation 2, we repeat the same approach using $\Gamma$ instead of $\Theta$.

To compute the thresholds $\tau_r$ and $\tau_c$, we first calculate $[v_{col}]_{ij} = \|\Theta_{\cdot i} - \Theta_{\cdot j}\|_2$ and stack this matrix to vector $v_{col}$; similarly we calculate $[v_{row}]_{ij} = \|\Theta_{i \cdot} - \Theta_{j \cdot}\|_2$ and stack to vector $v_{row}$. In the case of sparse linear regression, $\tau$ should be on the order of the noise \cite{meinshausen2009lasso}: $\tau \propto \sigma\sqrt{\log(p)/n}$, where $\sigma$ is typically estimated using the standard deviation of residuals. In \cite{chi2014convex} the authors set $\tau$ proportional to the standard deviation of $v_{row}$ or $v_{col}$.

However in our case, we have an additional regression loss term for estimating the parameters and hence there are two sources of randomness, the regression residual and the error in $v$. Taking these into account, we set $\tau_c = \frac 12 \Big[\sigma\sqrt{\log(p)/n} + std(v_{col})\Big]$ and $\tau_r = \frac 12 \Big[\sigma\sqrt{\log(p)/n} + std(v_{row})\Big]$. We set the multiplier to $\frac 1 2$, following the conservative suggestion in \cite{chi2014convex}.


\subsection{Specializing to Column- or Row-only Clustering (a.k.a. \textit{Uni-Clustering})}
\label{sec:uni}
Although formulations 1 and 2 have been developed for row-column bi-clustering, they can be easily specialized to clustering columns or rows alone, by respectively using only $\Omega_W(\Theta)$ or $\Omega_{\tilde{W}}(\Theta^T)$ in (\ref{eqn:form_1_MTL}), or using only $\Omega_W(\Gamma)$ or $\Omega_{\tilde{W}}(\Gamma^T)$ in (\ref{eqn:form_2_MTL}).


\section{SOLUTION PATH} \label{sec:solution_path}
 
Since we are able to obtain the entire solution path for the coefficients by varying the penalty multipliers, we provide an example of the solution paths for estimated $\Theta$ using formulation 1. The dataset is generated using the same procedure described in Section \ref{sec:sim_setup} except that we set $n=50$, $p=20$, and $k=15$. We use relatively small values for $p$ and $k$ since there will be a total of $pk$ solution paths to visualize. We will illustrate this only for formulation 1 since it is simpler. The solution paths for formulation 2 are provided in the supplementary material.

We first fix a reasonable $\lambda_1$ and vary $\lambda_2$ to get solution paths for all the coefficients. In our experiment, we chose $\lambda_1$ based on cross-validation as described in Section \ref{sec:weights_sparsity}. These paths are shown in Figure \ref{path_f1_vary_lambda2}. We can see that as $\lambda_2$ increases, the coefficients begin to merge and eventually for large enough $\lambda_2$ they are all equal. The solution paths are smooth in $\lambda_2$. Also note that the coefficient values are not monotonically increasing or decreasing, similar to the LASSO solution path \cite{tibshirani2011}.

Similarly, we fix $\lambda_2$ based on the cross-validation scheme described in Section \ref{sec:penalty_multiplier_tuning} and vary $\lambda_1$ to get solution paths for all the coefficients. This is shown in Figure \ref{path_f1_vary_lambda1}. It is well-known that the solution paths for LASSO are piecewise linear \cite{rosset2007piecewise}, when $L$ is least squares loss. Here, we see that the solution paths are not piecewise linear, but rather a smoothed version of it. This smoothness is imparted by the convex clustering regularization, the third term in (\ref{eqn:form_1_MTL}).

\begin{figure}[ht!]
\centering
\includegraphics[width=0.95\linewidth]{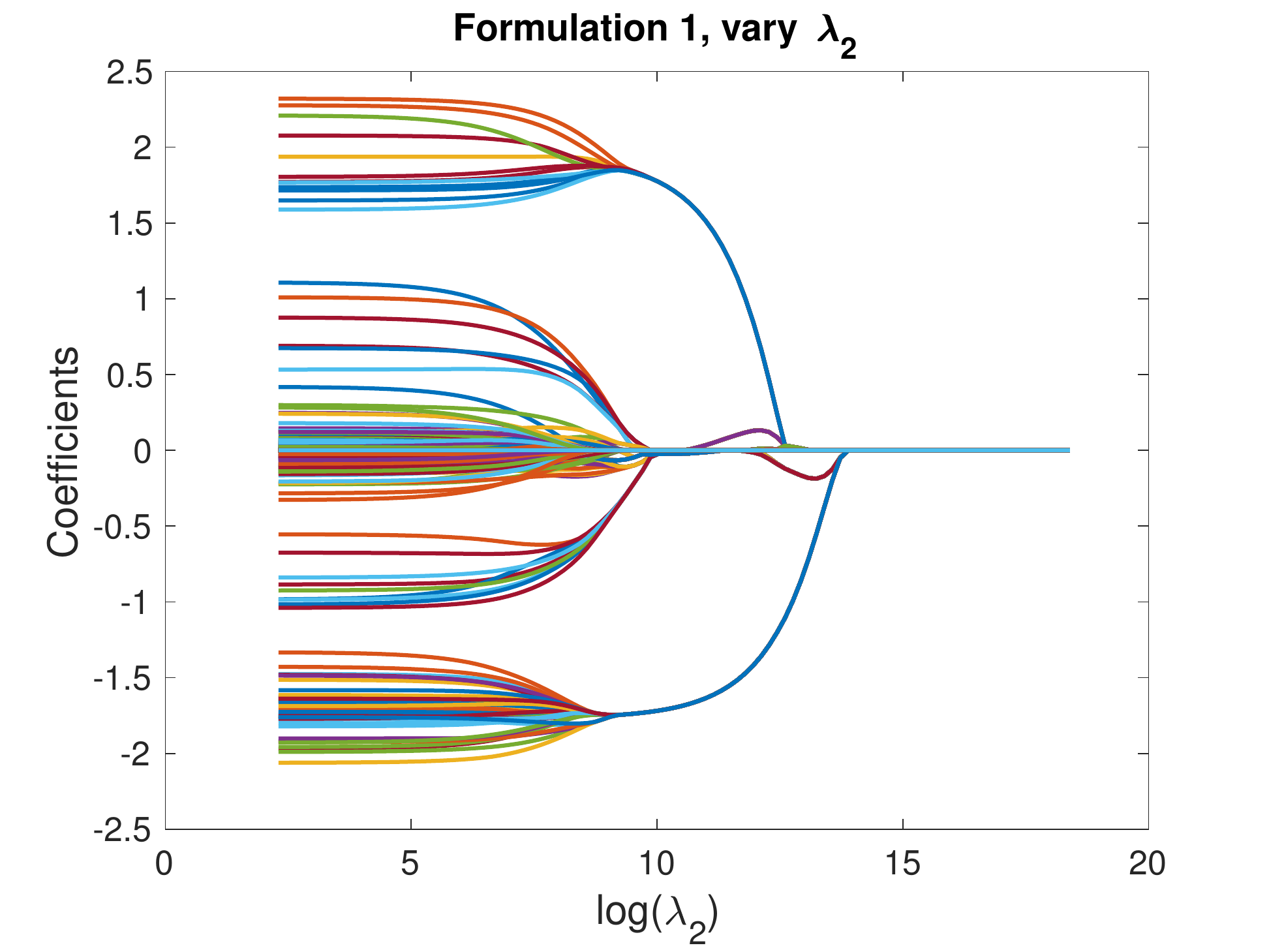}
\caption{Solution paths for formulation 1, fixing $\lambda_1$ and varying $\lambda_2$. Each line indicates a distinct coefficient.}
\label{path_f1_vary_lambda2}
\end{figure}

\begin{figure}[ht!]
\centering
\includegraphics[width=0.95\linewidth]{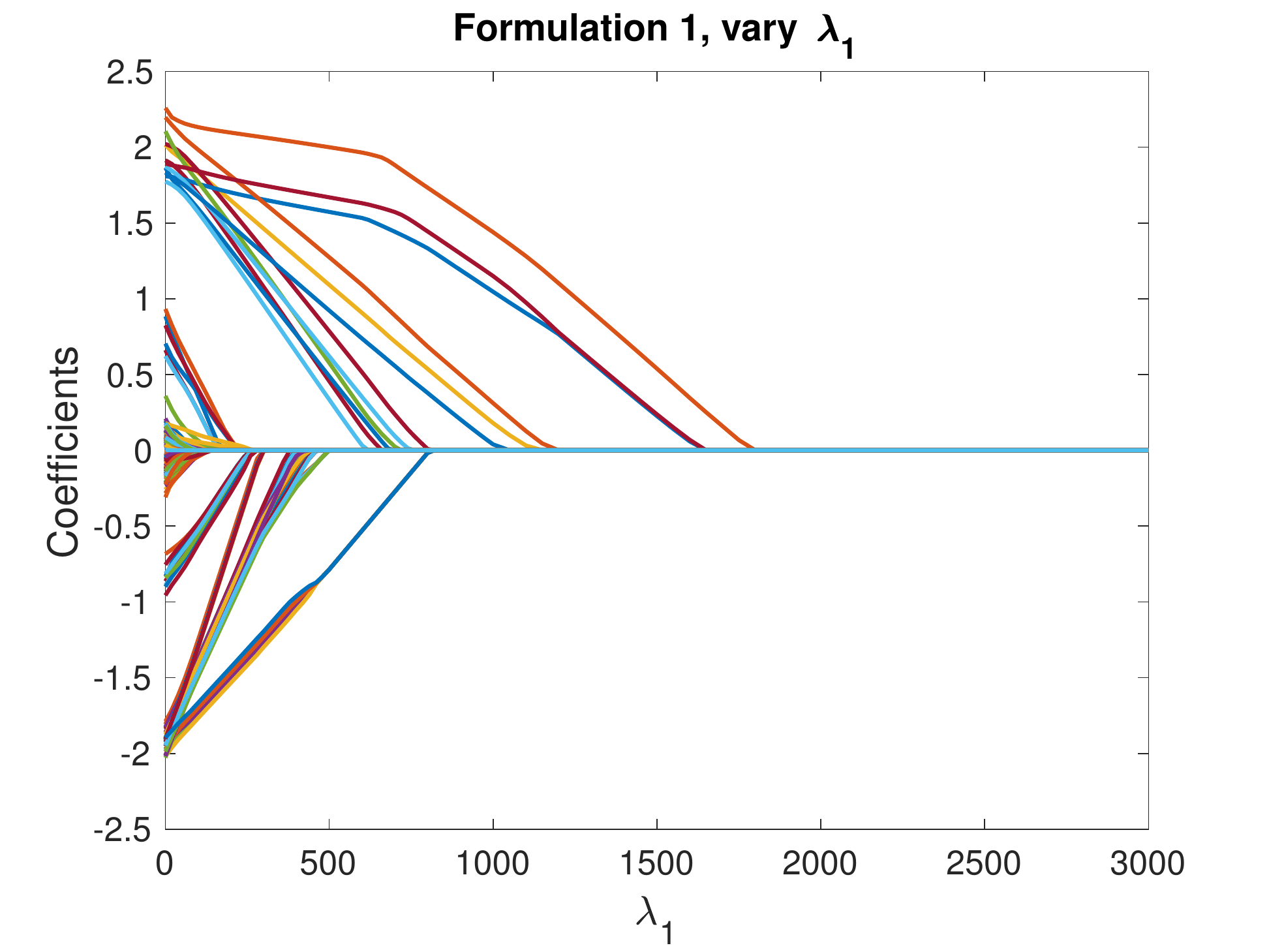}
\caption{Solution paths for formulation 1, fixing $\lambda_2$ and varying $\lambda_1$. Each line indicates a distinct coefficient.}
\label{path_f1_vary_lambda1}
\end{figure}

\section{SYNTHETIC DATA EXPERIMENTS} 
\label{sec:simulation}
We demonstrate our approach using experiments with synthetic data on the problem of multi-task learning. Detailed experiments on row- or column-alone clustering (\textit{uni-clustering}) are provided in supplementary materials.

We begin by describing the performance measures used to evaluate the clustering and estimation performance. 

\subsection{Performance Measures}

The estimation accuracy is measured by calculating the RMSE on an independent test set, and also the parameter recovery accuracy, $\|\hat \Theta_{est} - \Theta^*\| / \|\Theta^*\|$ where $\hat \Theta_{est}$ and $\Theta^*$ are the estimated and true coefficient matrices. Assessing the clustering quality can be hard. In this paper, 
we use the following three measures to evaluate the quality of clustering: the adjusted Rand index \cite{hubert1985comparing} (ARI), the F-1 score (F-1), and the Jaccard index (JI). The definitions of F-1 and JI are given in the supplementary material. For all these three measures, a value of 1 implies the best possible performance, and a value of 0 means that we are doing poorly. 


In order to compute ARI, F-1, and JI, we choose the value of the multiplier $\lambda_2$ in formulation 1, and  $\{ \lambda_2, \lambda_3\}$ in formulation 2 using the approach described in Section \ref{sec:penalty_multiplier_tuning}, and obtain the estimated clusterings. 

\subsection{Simulation Setup and Results}
\label{sec:sim_setup}
We focus on multi-task regression: $ Y = X \Theta^* + E$ with $e_{ij} \sim N(0, \sigma^2)$. 
All the entries of design matrix $X$ are generated as independent standard normal. 
The true regression parameter $\Theta^*$ has a bi-cluster (checkerboard) structure. To simulate sparsity, we set the coefficients within many of the blocks in the checkerboard to $0$. For the non-zero blocks, we follow the generative model recommended in \cite{chi2014convex}: the coefficients within each cluster are generated as $\theta_{ij} = \mu_{rc} + \epsilon_{ij}$ with $\epsilon_{ij} \sim N(0, \sigma_\epsilon^2)$ to make them close but not identical, where $\mu_{rc}$ is the mean of the cluster defined by the $r^{th}$ row partition and $c^{th}$ column partition. We set $n = 200$, $p = 500$, and $k = 250$ in our experiment. For the non-zero blocks, we set $\mu_{rc} \sim \text{Uniform}\{-2, -1, 1, 2\}$ and set $\sigma_\epsilon = 0.25$. We try the low-noise setting ($\sigma = 1.5$), where it is relatively easy to estimate the clusters, and the high-noise setting ($\sigma = 3$), where it is harder to obtain them.

We compare our formulation 1 and formulation 2 with a 2-step \emph{estimate-then-cluster} approach: (a) Estimate $\hat \Theta$ first using LASSO, and (b) perform convex bi-clustering on $\hat \Theta$. $\hat \Theta$ is estimated by solving \eqref{eqn:Lasso} while selecting the best $\lambda_1$ as discussed in Section \ref{sec:weights_sparsity}, and the convex bi-clustering step is implemented using COBRA algorithm in \cite{chi2014convex}. Our \textit{baseline} clustering performance is the best of the following: (a) letting each coefficient be its own group, and (b) imposing a single group for all coefficients. 



The average clustering quality results on 50 replicates are shown in Table \ref{high_snr} and Table \ref{low_snr} for low and high noise settings, respectively. In both tables, the first, second, and third blocks correspond to performances of row, column and row-column bi-clusterings, respectively. We optimize only for bi-clusterings, but the row and the column clusterings are obtained as by-products. Note that this could lead to different results compared to directly performing uni-clustering.

The RMSEs evaluated on the test set and the parameter recovery accuracy are provided in Table \ref{estimation_accuracy_high} and Table \ref{estimation_accuracy_low}. Most performance measures are reported in the format $mean \pm std. dev.$ 


From Table \ref{high_snr} and Table \ref{low_snr} we see that both our formulation 1 and 2 give better results on row clustering, column clusterings, and row-column bi-clustering compared to the 2-step procedure. Moreover, the clustering results given by our formulations are more stable, with lesser spread in performance.


We compare the RMSE and the parameter recovery accuracy of the proposed formulations with other approaches and report the results in Table \ref{estimation_accuracy_high} and Table \ref{estimation_accuracy_low}. The oracle RMSE (with $\Theta$ known) is $1.5$ for Table \ref{estimation_accuracy_high} and $3.0$ for Table \ref{estimation_accuracy_low}, and we can see that the proposed methods provide improvements over the others.  We also observe improvements in the parameter recovery accuracy.

The performance boost obtained with high noise is much higher compared to that with low noise. This makes sense because when noise level is low, the estimation step in the 2-step approach is more accurate and the error propagated into the clustering step is relatively small. However at high noise levels, the estimation can be inaccurate. This estimation error propagates into the clustering step and makes the clustering result of 2-step approach unreliable. However, our formulations are able to jointly do the estimation and clustering, and hence have more reliable and stable results.





\begin{table}[ht]
\begin{small}
\setlength{\tabcolsep}{4.5pt}
\caption{Performance for low noise ($\sigma = 1.5$) setting. First, second, and third blocks correspond to row clustering, column clustering, and row-column bi-clustering.}
\label{high_snr}
\begin{center}
\begin{tabular}{c|cccc}
  & Baseline & 2-step & Form1 & Form2\\  \hline
ARI    &   0 &   0.679$\pm$0.157  &  0.869$\pm$0.069  &  0.900$\pm$0.046 \\
F-1 & 0.446  &  0.757$\pm$0.128  &  0.907$\pm$0.052  &  0.931$\pm$0.022 \\
JI  & 0.287  &  0.625$\pm$0.161  &  0.834$\pm$0.081  &  0.871$\pm$0.042 \\
%
%
\hline
  ARI     &   0  &  0.877$\pm$0.043  &  0.914$\pm$0.020  &  0.915$\pm$0.013\\
F-1 & 0.446 &  0.908$\pm$0.037 &   0.933$\pm$0.023   & 0.934$\pm$0.012\\
JI  & 0.287 &   0.847$\pm$0.048  &  0.876$\pm$0.031  &  0.887$\pm$0.025\\
%
\hline
ARI   &    0  &  0.708$\pm$0.118  &  0.841$\pm$0.059  &  0.863$\pm$0.035\\
F-1 & 0.172  &  0.734$\pm$0.110  &  0.857$\pm$0.052  &  0.877$\pm$0.026 \\
JI & 0.094  &  0.591$\pm$0.134  &  0.753$\pm$0.077  &  0.781$\pm$0.035
\end{tabular}
\end{center}
\end{small}
\end{table}

\begin{table}[ht]
\begin{small}
\setlength{\tabcolsep}{2pt}
\caption{RMSE and parameter recovery accuracy of the estimation schemes of low noise ($\sigma = 1.5$) setting.}
\label{estimation_accuracy_high}
\begin{center}
\begin{tabular}{c|cccc}
 & Lasso&2-step & Form1 & Form2 \\\hline
RMSE & 1.627$\pm$0.02 &   1.622$\pm$0.02 &   1.613$\pm$0.02  &  1.612$\pm$0.02 \\ 
Rec. acc. & 0.234$\pm$0.03  &  0.231$\pm$0.03   & 0.223$\pm$0.03 &   0.222$\pm$0.03  \\ 
\end{tabular}
\end{center}
\end{small}
\end{table}


\begin{table}[ht]
\begin{small}
\setlength{\tabcolsep}{4.5pt}
\caption{Performance for high noise ($\sigma = 3$) setting. First, second, and third blocks correspond to row clustering, column clustering, and row-column bi-clustering.}
\label{low_snr}
\begin{center}
\begin{tabular}{c|cccc}
  & Baseline & 2-step & Form1 & Form2\\  \hline
ARI    &   0 &   0.577$\pm$0.163  &  0.803$\pm$0.104  &  0.804$\pm$0.096 \\
F-1 & 0.446  &  0.674$\pm$0.138  &  0.874$\pm$0.093  &  0.874$\pm$0.075 \\
JI  & 0.287  &  0.525$\pm$0.159  &  0.793$\pm$0.097  &  0.792$\pm$0.098 \\
\hline
  ARI     &   0  &  0.734$\pm$0.132  &  0.905$\pm$0.077  &  0.905$\pm$0.046\\
F-1 & 0.446 &  0.799$\pm$0.107 &   0.924$\pm$0.054   & 0.933$\pm$0.039\\
JI  & 0.287 &   0.689$\pm$0.120  &  0.872$\pm$0.078  &  0.867$\pm$0.065\\
\hline
ARI   &    0  &  0.555$\pm$0.187  &  0.801$\pm$0.125  &  0.812$\pm$0.105\\
F-1 & 0.172  &  0.586$\pm$0.152  &  0.824$\pm$0.104  &  0.821$\pm$0.086 \\
JI & 0.094  &  0.437$\pm$0.179  &  0.714$\pm$0.118  &  0.713$\pm$0.104
\end{tabular}
\end{center}
\end{small}
\end{table}

\begin{table}[ht]
\begin{small}
\setlength{\tabcolsep}{2pt}
\caption{RMSE and parameter recovery accuracy of the estimation schemes of high noise ($\sigma = 3$) setting.}
\label{estimation_accuracy_low}
\begin{center}
\begin{tabular}{c|cccc}
 & Lasso&2-step & Form1 & Form2 \\\hline
RMSE & 3.34$\pm$0.02 &   3.30$\pm$0.02 &   3.23$\pm$0.02  &  3.16$\pm$0.02 \\ 
Rec. acc. & 0.364$\pm$0.06  &  0.362$\pm$0.06   & 0.327$\pm$0.05 &   0.325$\pm$0.06  \\ 
\end{tabular}
\end{center}
\end{small}
\end{table}


\section{REAL DATA EXPERIMENTS} \label{sec:real_data}

We demonstrate the proposed approaches using real datasets obtained from experiments with sorghum crops \cite{tuinstra2016automated}. Accurate phenotyping of different crop varieties is a crucial yet traditionally a time-consuming step in crop breeding, requiring manual survey of hundreds of plant varieties for the traits of interest. Typical workflow of recent automated, high-throughput remote sensing systems for trait development and GWAS is shown in Figure \ref{appl_system}. We consider two specific problems from this pipeline: (a) predictive modeling of plant traits using features from remote sensed data (Section \ref{trait_prediction_expt}), (b) GWAS using the reference traits (Section \ref{GWAS_reference}). Additional experiments are provided in the supplementary material.

\begin{figure}[htbp]
\begin{center}
\includegraphics[width=0.48\textwidth]{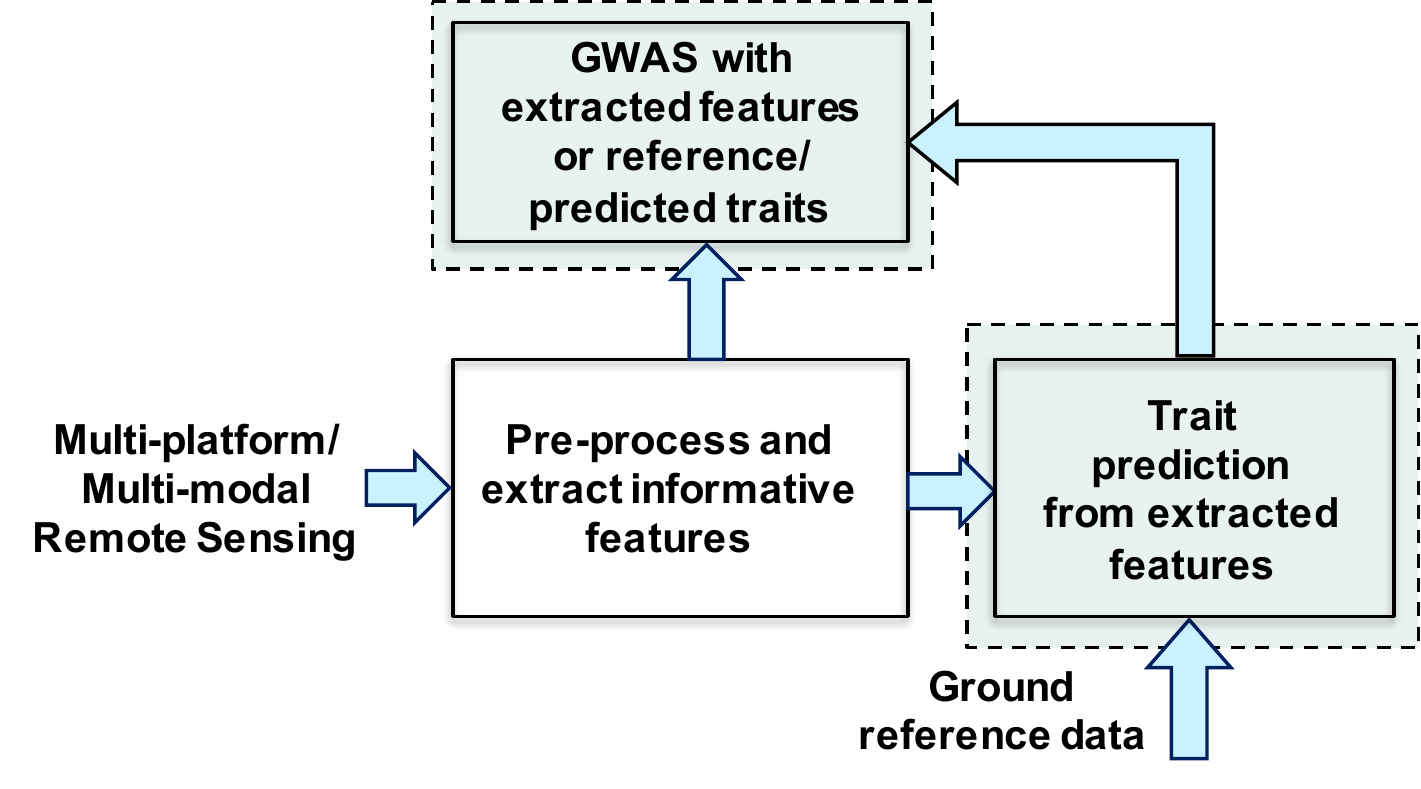}
\caption{Context of our real data experiments. We apply our proposed approaches for mapping from input remote sensing features to output plant traits (Section \ref{trait_prediction_expt}), and GWAS with reference traits (Section \ref{GWAS_reference}).}
\label{appl_system}
\end{center}
\end{figure}

\subsection{Phenotypic Trait Prediction from Remote Sensed Data}
\label{trait_prediction_expt}

\begin{figure}[!ht]
\begin{center}
\centerline{\includegraphics[width=0.99\columnwidth]{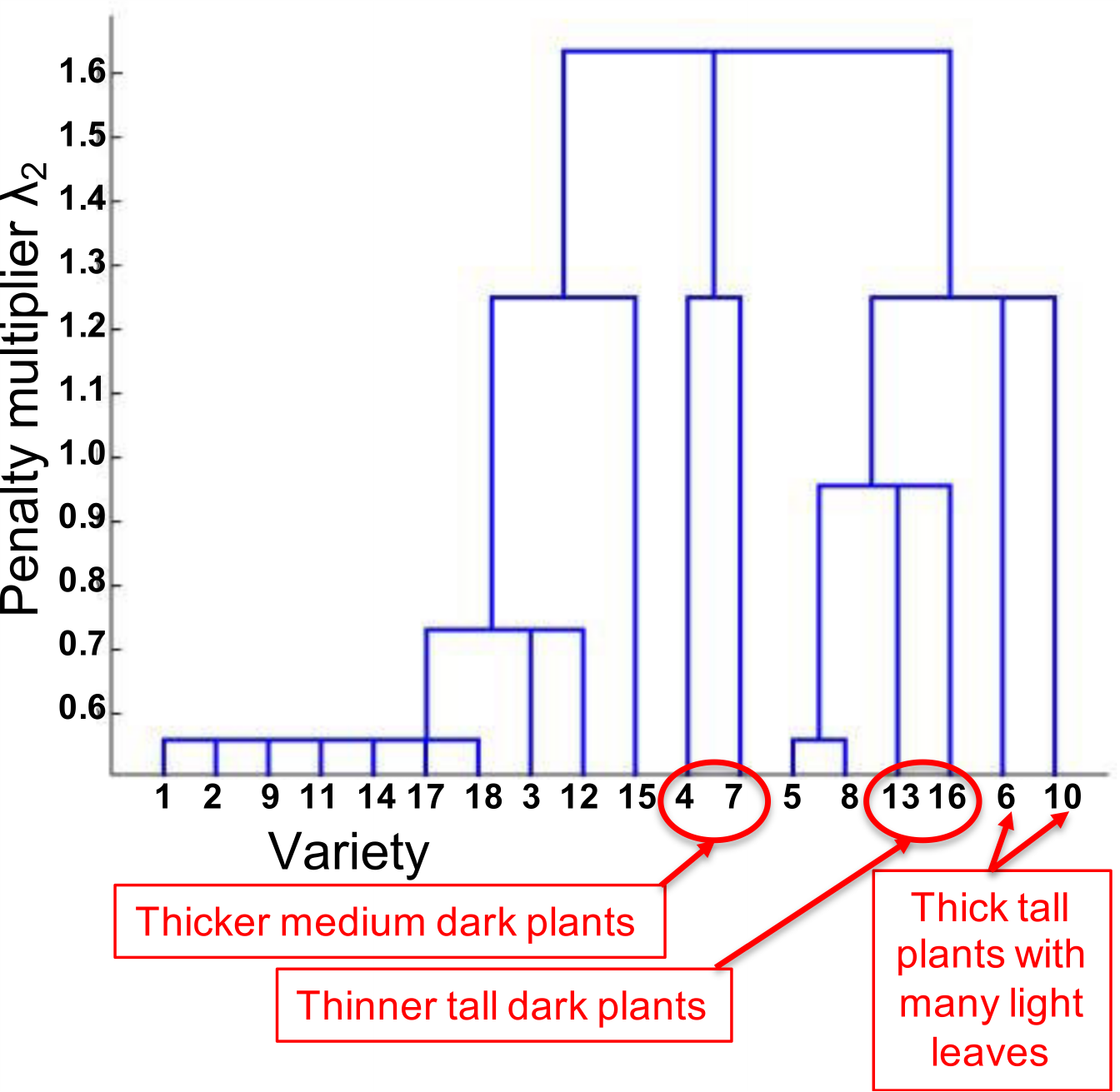}}
\caption{Tree structure of tasks (varieties) inferred using our approach for plant height.}
\label{Ph1}
\end{center}
\end{figure}

The experimental data was obtained from $18$ sorghum varieties planted in $6$ replicate plot locations, and we considered three different traits: plant height, stalk diameter, and stalk volume. We report results only for plant height here and the results for other traits, along with $18$ variety names are given in the supplementary material.

From the RGB and hyperspectral images of each plot, we extract features of length $206$. Hence $n=6$, $p=206$, and the number of tasks $k=18$, for each trait considered. The presence of multiple varieties with replicates much smaller in number than predictors poses a major challenge: building separate models for each variety is unrealistic, while a single model does not fit all. This is where our proposed simultaneous estimation and clustering approach provides the flexibility to share information among tasks that leads to learning at the requisite level of robustness. Note that here we use the column-only clustering variant of formulation 1.

The dendrogram for task clusters obtained by sweeping the penalty multiplier $\lambda_2$ is given in Figure \ref{Ph1}. This provides some interesting insights from a plant science perspective. As highlighted in Figure~\ref{Ph1}, the predictive models (columns of $\Theta$) for thicker medium dark plants are grouped together. Similar grouping is seen for thinner tall dark plants, and thick tall plants with many light leaves.

To compute RMSE, we perform  6-folds CV where each fold consists of at least one example from each variety. As we only have $n=6$ samples per variety (i.e. per task), it is unrealistic to learn separate models for each variety. For each CV split, we first learn a grouping using one of the compared methods, treat all the samples within a group as i.i.d, and estimate their regression coefficients using Lasso. The methods compared with our approach include: (a) \textit{single model}, which learns a single predictive model using Lasso, treating all the varieties as i.i.d., (b) \textit{No group multitask learning}, which learns a traditional multitask model using Group Lasso where each variety forms a separate group, and (c) Kang \textit{et al.} \cite{kang2011learning}, which uses a mixed integer program to learn shared feature representations among tasks, while simultaneously determining ``with whom'' each task should share. Results reported in Table~\ref{RMSEphen}, indicate the superior quality of our groupings in terms of improved predictive accuracy.

\begin{table}[h]
\caption{RMSE for plant height prediction.}
\label{RMSEphen}
\vskip 0.15in
\begin{center}
\begin{small}
\begin{tabular}{lccc}
\hline
Method & RMSE \\
\hline
Single model &  44.39$\pm$6.55 \\
No group multitask learning &  36.94$\pm$6.10  \\
Kang \textit{et al.} & 37.55$\pm$7.60 \\
Proposed &  \bf{33.31$\pm$5.10} \\
\hline
\end{tabular}
\end{small}
\end{center}
\vskip -0.1in
\end{table}

\begin{figure}[!ht]
\begin{center}
\centerline{\includegraphics[width=1.1\columnwidth]{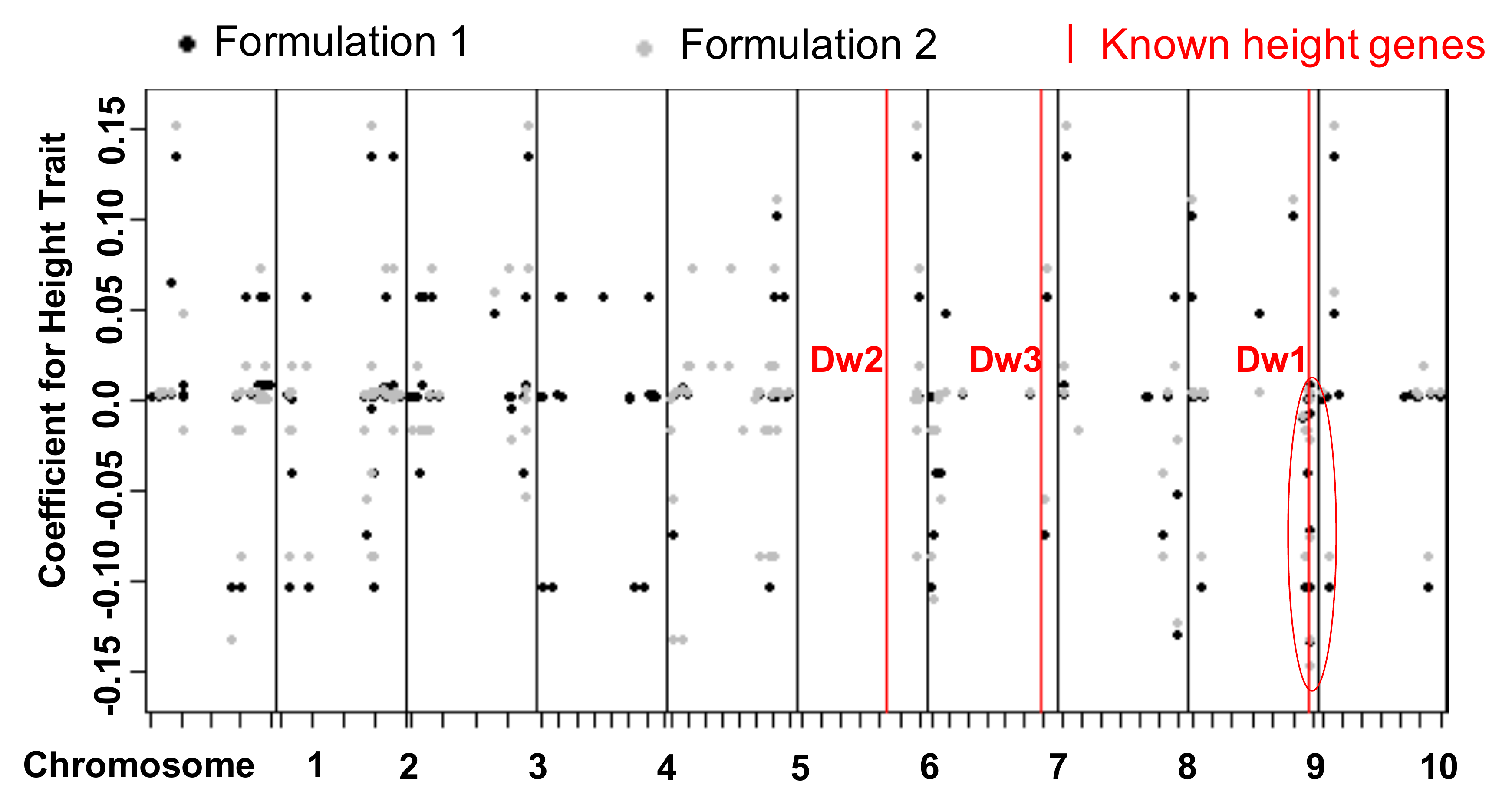}}
\caption{Distribution of coefficients for height traits for all SNPs}
\label{chr_figure}
\end{center}
\end{figure}

\begin{figure}[ht!]
\begin{center}
\begin{minipage}[t]{0.49\linewidth}
\centering
\includegraphics[width=0.99\textwidth]{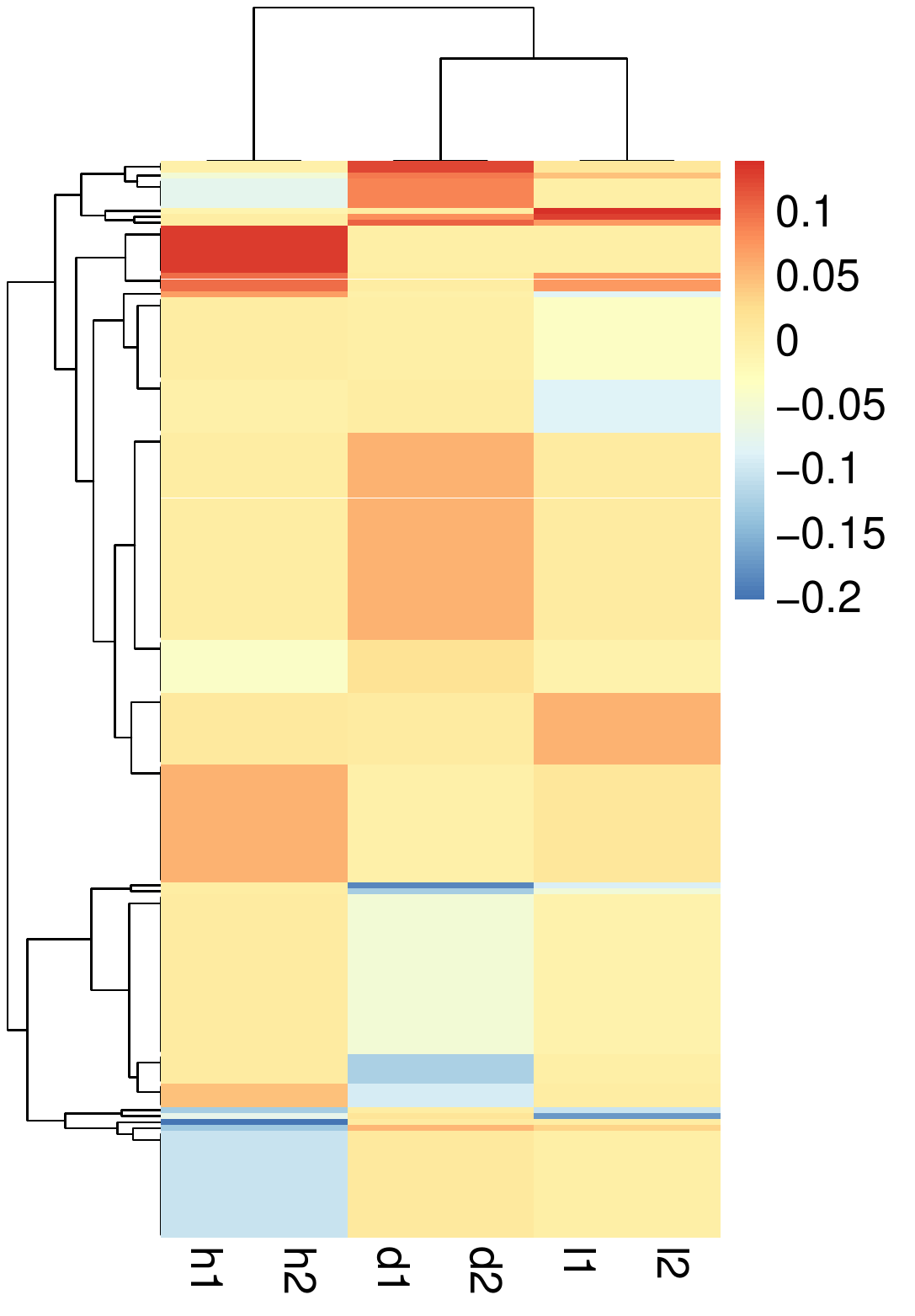}
\end{minipage}
\begin{minipage}[t]{0.49\linewidth}
\centering
\includegraphics[width=0.99\textwidth]{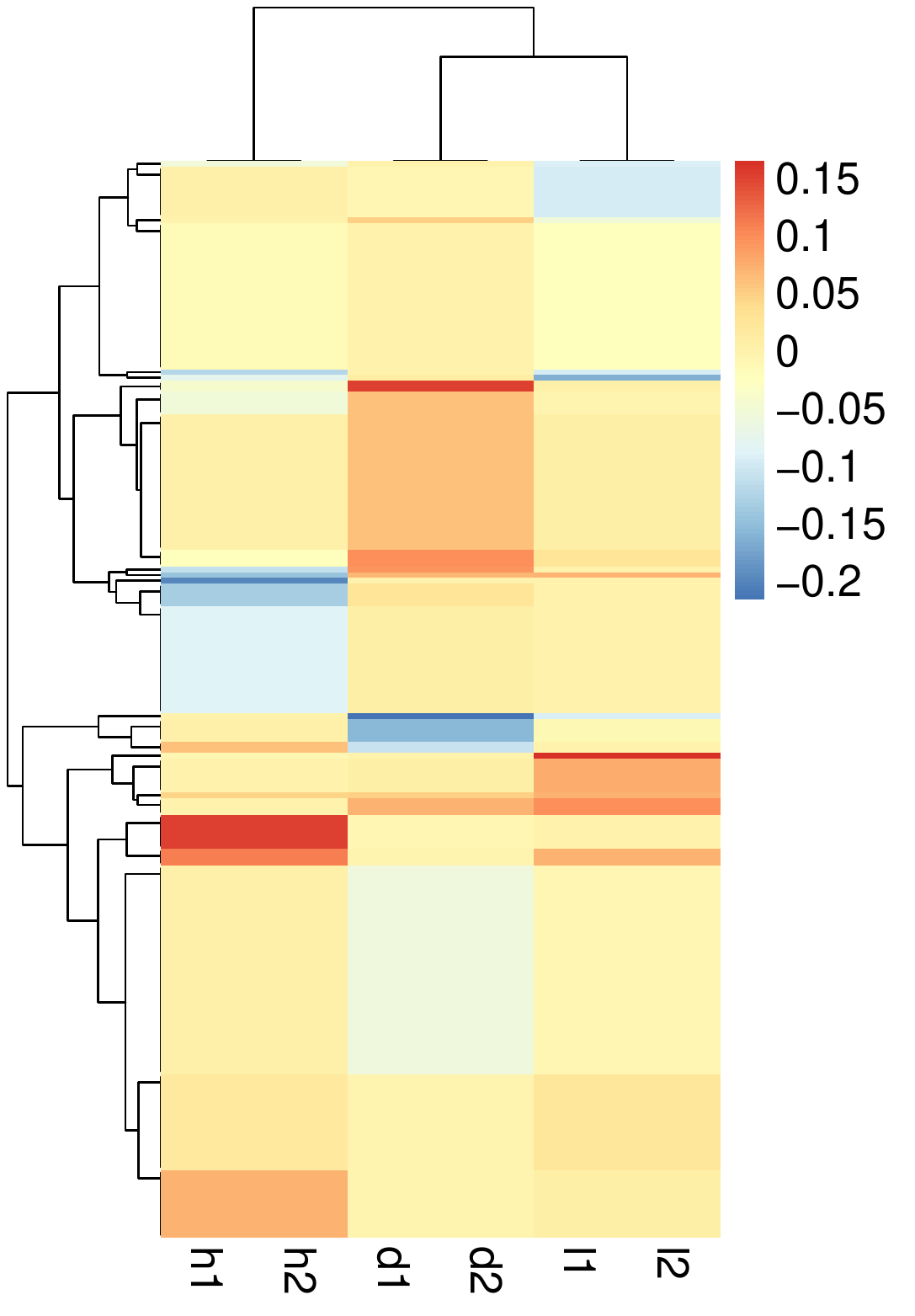}
\end{minipage}
\end{center}
\caption{Smoothed coefficient matrix obtained from formulations 1 (left) and 2 (right), revealing the bi-clustering structure.}
\label{GWAS_bicluster_f1_f2}
\end{figure}

%

\subsection{Multi-Response GWAS}
\label{GWAS_reference}
We apply our approach in a multi-response Genome-Wide Association Study (GWAS). While traditional GWAS focuses on associations to single phenotypes, we would like to automatically learn the grouping structure between the phenotypes as well as the features (columns and rows of $\Theta$) using our proposed method. We use the proposed formulations 1 and 2 (bi-clustering variant) in this experiment.

The design matrix $X$ consisted of SNPs of Sorghum varieties. We consider $n=911$ varieties and over $80,000$ SNPs. We remove duplicate SNPs and also SNPs that do not have significantly high correlation to at least one response variable. Finally, we end up considering $p=2,937$ SNPs. The output data $Y$ contains the following $k=6$ response variables (columns) for all the $n$ varieties collected by hand measurements:
\begin{enumerate}
\item \emph{Height to panicle} (h1): The height of the plant up to the panicle of the sorghum plant.
\item \emph{Height to top collar} (h2): The height of the plant up to the top most leaf collar.
\item \emph{Diameter top collar} (d1): The diameter of the stem at the top most leaf collar.
\item \emph{Diameter at 5 cm from base} (d2): The diameter of the stem at 5 cm from the base of the plant.
\item \emph{Leaf collar count} (l1): The number of leaf collars on the plant.
\item  \emph{Green leaf count} (l2): The total number of green leaves. This will be less than l1 since some leaves may have senesced and will not be green anymore.
\end{enumerate} Each trait can be an average of measurements from up to $4$ plants, in every variety.

\begin{table}[t]
\caption{Comparison of test RMSE on the multi-response GWAS dataset.} 
\label{test_RMSE}
\begin{center}
\begin{tabular}{c|cccc}
 & Lasso  & 2-step & Form1 & Form2 \\\hline
RMSE & 2.181 & 2.206 & 2.105 & 2.119
\end{tabular}
\end{center}
\end{table}

We split our data set into 3 parts: 70\% training, 15\% validation, and 15\% test. We estimate the coefficient matrices by optimizing our formulations on the training set, select the tuning parameters based on the validation set (Sections \ref{sec:penalty_multiplier_tuning}, \ref{sec:biclus_thresh}), and then calculate the RMSE on the test set. Table \ref{test_RMSE} shows the RMSE on test set. The coefficient matrix given by our formulations are visualized in Figure \ref{GWAS_bicluster_f1_f2}.  To make the figure easier to interpret, we exclude the rows with all zero coefficients and take the average over the coefficients within each bi-cluster. The light yellow areas in the figures are zero coefficients; red and blue areas are positive and negative coefficients, respectively. The rows and columns are reordered to best show the checkerboard patterns. We wish to emphasize again that this checkerboard pattern is automatically discovered using our proposed procedures, and is not evidently present in the data.


The two formulations reveal similar bi-clustering patterns up to reordering. For column clusters, the plant height tasks (h1 and h2), the stem diameter tasks (d1 and d2), and the leaf tasks (l1 and l2) group together. Also, the stem diameter and leaf tasks are more related to each other compared to the height tasks. The bi-clustering patterns reveal the group of SNPs that influence similar phenotypic traits. Coefficients for height features in the GWAS (Figure \ref{chr_figure}) study show SNPs with strong effects coinciding with locations of Dwarf 3 \cite{multani2003loss} and especially Dwarf 1 \cite{hilley2016identification} genes known to control plant height that are segregating and significant in the population. The lack of any effect at the Dwarf 2 \cite{hilley2017sorghum} locus supports previous work indicating that this gene is not a strong contributing factor in this population.

We also estimate the RMSE of the proposed formulations and compare it with the RMSE provided by a simple Lasso model and 2-step procedure. This is shown in Table \ref{test_RMSE}. We see that the RMSE of our formulations are slightly less than that of the Lasso and 2-step procedure. Hence, for similar estimation performance, we are able to discover additional interesting structure in the input-output relationship using our proposed methods.



\section{CONCLUDING REMARKS} \label{sec:conclusion}
In this paper we introduced and studied formulations for joint estimation and clustering (row or column or both) of the parameter matrix in multi-response models. By design, our formulations imply that coefficients belonging to the same (bi-)cluster are close to one another. By incorporating different notions of closeness between the coefficients, we can tremendously increase the scope of applications in which similar formulations can be used. Some future applications could include sparse subspace clustering and community detection.

Recently there has been a lot of research on non-convex optimization formulations, both from theoretical and empirical perspectives \cite{wang2014nonconvex, Yu2017AnIM}. It would be of interest to see the performance of our formulations on non-convex loss functions. Another extension would be to construct confidence intervals and perform hypothesis testing for the coefficients in each cluster. There are two kinds of bias in our formulations: (a) shrinkage bias due to the $\ell_1$ regularization, and (b) bias due to the bi-clustering regularization, because of the fact that it forces coefficients to be close to each other. Many de-biasing methods have been proposed to handle the first type of bias \cite{van2014asymptotically, yu2016statistical, javanmard2014confidence,ning2017general}, while de-biasing methods for the second type of bias is a potential research area.


\section*{Acknowledgment}
The information, data, or work presented herein was funded in part by the Advanced Research Projects Agency-Energy (ARPA-E), U.S. Department of Energy, under Award Number {DE-AR0000593}. The views and opinions of authors expressed herein do not necessarily state or reflect those of the United States Government or any agency thereof.
%
{\footnotesize
\bibliographystyle{plain}
\bibliography{paper}

\begin{thebibliography}{10}

\bibitem{beck2015convergence}
Amir Beck.
\newblock On the convergence of alternating minimization for convex programming
  with applications to iteratively reweighted least squares and decomposition
  schemes.
\newblock {\em SIAM Journal on Optimization}, 25(1):185--209, 2015.

\bibitem{borchani2015survey}
Hanen Borchani, Gherardo Varando, Concha Bielza, and Pedro Larra{\~n}aga.
\newblock A survey on multi-output regression.
\newblock {\em Wiley Interdisciplinary Reviews: Data Mining and Knowledge
  Discovery}, 5(5):216--233, 2015.

\bibitem{caruana1998multitask}
Rich Caruana.
\newblock Multitask learning.
\newblock In {\em Learning to learn}, pages 95--133. Springer, 1998.

\bibitem{chi2014convex}
Eric~C Chi, Genevera~I Allen, and Richard~G Baraniuk.
\newblock Convex biclustering.
\newblock {\em arXiv preprint arXiv:1408.0856}, 2014.

\bibitem{chi2015splitting}
Eric~C Chi and Kenneth Lange.
\newblock Splitting methods for convex clustering.
\newblock {\em Journal of Computational and Graphical Statistics},
  24(4):994--1013, 2015.

\bibitem{combettes2008proximal}
Patrick~L Combettes and Jean-Christophe Pesquet.
\newblock A proximal decomposition method for solving convex variational
  inverse problems.
\newblock {\em Inverse problems}, 24(6):065014, 2008.

\bibitem{hallac2015network}
David Hallac, Jure Leskovec, and Stephen Boyd.
\newblock Network lasso: Clustering and optimization in large graphs.
\newblock In {\em Proceedings of the 21th ACM SIGKDD international conference
  on knowledge discovery and data mining}, pages 387--396. ACM, 2015.

\bibitem{mcwilliams2014loco}
Christina Heinze, Brian McWilliams, Nicolai Meinshausen, and Gabriel
  Krummenacher.
\newblock Loco: Distributing ridge regression with random projections.
\newblock {\em arXiv preprint arXiv:1406.3469}, 2014.

\bibitem{hilley2016identification}
Josie Hilley, Sandra Truong, Sara Olson, Daryl Morishige, and John Mullet.
\newblock Identification of \textit{Dw1}, a regulator of sorghum stem internode
  length.
\newblock {\em PLoS One}, 11(3):e0151271, 2016.

\bibitem{hilley2017sorghum}
Josie~L Hilley, Brock~D Weers, Sandra~K Truong, Ryan~F McCormick, Ashley~J
  Mattison, Brian~A McKinley, Daryl~T Morishige, and John~E Mullet.
\newblock Sorghum \textit{Dw2} encodes a protein kinase regulator of stem
  internode length.
\newblock {\em Scientific Reports}, 7(1):4616, 2017.

\bibitem{hubert1985comparing}
Lawrence Hubert and Phipps Arabie.
\newblock Comparing partitions.
\newblock {\em Journal of classification}, 2(1):193--218, 1985.

\bibitem{jacob2009clustered}
Laurent Jacob, Jean-philippe Vert, and Francis~R Bach.
\newblock Clustered multi-task learning: A convex formulation.
\newblock In {\em Advances in Neural Information Processing Systems}, pages
  745--752, 2009.

\bibitem{jalali2010dirty}
Ali Jalali, Sujay Sanghavi, Chao Ruan, and Pradeep~K Ravikumar.
\newblock A dirty model for multi-task learning.
\newblock In {\em Advances in Neural Information Processing Systems}, pages
  964--972, 2010.

\bibitem{javanmard2014confidence}
Adel Javanmard and Andrea Montanari.
\newblock Confidence intervals and hypothesis testing for high-dimensional
  regression.
\newblock {\em The Journal of Machine Learning Research}, 15(1):2869--2909,
  2014.

\bibitem{kang2011learning}
Zhuoliang Kang, Kristen Grauman, and Fei Sha.
\newblock Learning with whom to share in multi-task feature learning.
\newblock In {\em Proceedings of the 28th International Conference on Machine
  Learning (ICML-11)}, pages 521--528, 2011.

\bibitem{kim2010tree}
Seyoung Kim and Eric~P. Xing.
\newblock Tree-guided group lasso for multi-task regression with structured
  sparsity.
\newblock In {\em Proceedings of the 27th International Conference on Machine
  Learning (ICML-10)}, pages 543--550, 2010.

\bibitem{kumar2012learning}
Abhishek Kumar and Hal Daume~III.
\newblock Learning task grouping and overlap in multi-task learning.
\newblock {\em arXiv preprint arXiv:1206.6417}, 2012.

\bibitem{lu2014fast}
Yichao Lu and Dean~P Foster.
\newblock Fast ridge regression with randomized principal component analysis
  and gradient descent.
\newblock {\em arXiv preprint arXiv:1405.3952}, 2014.

\bibitem{meinshausen2009lasso}
Nicolai Meinshausen and Bin Yu.
\newblock Lasso-type recovery of sparse representations for high-dimensional
  data.
\newblock {\em The Annals of Statistics}, pages 246--270, 2009.

\bibitem{multani2003loss}
Dilbag~S Multani, Steven~P Briggs, Mark~A Chamberlin, Joshua~J Blakeslee,
  Angus~S Murphy, and Gurmukh~S Johal.
\newblock Loss of an {MDR} transporter in compact stalks of maize \textit{br2}
  and sorghum \textit{dw3} mutants.
\newblock {\em Science}, 302(5642):81--84, 2003.

\bibitem{ning2017general}
Yang Ning and Han Liu.
\newblock A general theory of hypothesis tests and confidence regions for
  sparse high dimensional models.
\newblock {\em The Annals of Statistics}, 45(1):158--195, 2017.

\bibitem{obozinski2006multi}
Guillaume Obozinski, Ben Taskar, and Michael Jordan.
\newblock Multi-task feature selection.
\newblock {\em Statistics Department, UC Berkeley, Tech. Rep}, 2, 2006.

\bibitem{ramamurthy2016predictive}
Karthikeyan~Natesan Ramamurthy, Zhou Zhang, Addie Thompson, Fangning He, Melba
  Crawford, Ayman Habib, Clifford Weil, and Mitchell Tuinstra.
\newblock Predictive modeling of sorghum phenotypes with airborne image
  features.
\newblock In {\em Proc. of KDD Workshop on Data Science for Food, Energy, and
  Water}, 2016.

\bibitem{rosset2007piecewise}
Saharon Rosset and Ji~Zhu.
\newblock Piecewise linear regularized solution paths.
\newblock {\em The Annals of Statistics}, pages 1012--1030, 2007.

\bibitem{schifano2013genome}
Elizabeth~D Schifano, Lin Li, David~C Christiani, and Xihong Lin.
\newblock Genome-wide association analysis for multiple continuous secondary
  phenotypes.
\newblock {\em The American Journal of Human Genetics}, 92(5):744--759, 2013.

\bibitem{tibshirani2011}
Ryan~J. Tibshirani and Jonathan Taylor.
\newblock The solution path of the generalized lasso.
\newblock {\em Ann. Statist.}, 39(3):1335--1371, 06 2011.

\bibitem{tuinstra2016automated}
Mitch Tuinstra, Cliff Weil, Addie Thompson, Chris Boomsma, Melba Crawford,
  Ayman Habib, Edward Delp, Keith Cherkauer, Larry Biehl, Naoki Abe, Meghana
  Kshirsagar, Aurelie Lozano, Karthikeyan~Natesan Ramamurthy, Peder Olsen, and
  Eunho Yang.
\newblock Automated sorghum phenotyping and trait development platform.
\newblock In {\em Proc. of KDD Workshop on Data Science for Food, Energy, and
  Water}, 2016.

\bibitem{van2014asymptotically}
Sara Van~de Geer, Peter B{\"u}hlmann, Ya’acov Ritov, Ruben Dezeure, et~al.
\newblock On asymptotically optimal confidence regions and tests for
  high-dimensional models.
\newblock {\em The Annals of Statistics}, 42(3):1166--1202, 2014.

\bibitem{wang2014nonconvex}
Zhaoran Wang, Huanran Lu, and Han Liu.
\newblock Nonconvex statistical optimization: Minimax-optimal sparse pca in
  polynomial time.
\newblock {\em arXiv preprint arXiv:1408.5352}, 2014.

\bibitem{Yu2017AnIM}
Ming Yu, Varun Gupta, and Mladen Kolar.
\newblock An influence-receptivity model for topic based information cascades.
\newblock {\em 2017 IEEE International Conference on Data Mining (ICDM)}, pages
  1141--1146, 2017.

\bibitem{yu2016statistical}
Ming Yu, Mladen Kolar, and Varun Gupta.
\newblock Statistical inference for pairwise graphical models using score
  matching.
\newblock In {\em Advances in Neural Information Processing Systems}, pages
  2829--2837, 2016.

\bibitem{Yu2017multitask}
Ming {Yu}, Addie~M. {Thompson}, Karthikeyan {Natesan Ramamurthy}, Eunho {Yang},
  and Aurelie~C. {Lozano}.
\newblock {Multitask Learning using Task Clustering with Applications to
  Predictive Modeling and GWAS of Plant Varieties}.
\newblock {\em ArXiv e-prints}, October 2017.

\bibitem{yu2018recovery}
Ming Yu, Zhaoran Wang, Varun Gupta, and Mladen Kolar.
\newblock Recovery of simultaneous low rank and two-way sparse coefficient
  matrices, a nonconvex approach.
\newblock {\em arXiv preprint arXiv:1802.06967}, 2018.

\end{thebibliography}
}

\appendix
\section*{Supplementary Material} \label{sec:appendix}

This material provides additional details to support the main paper. In particular, we provide additional solution paths, define the clustering error measures used, provide detailed optimization steps and convergence proofs for formulations 1 and 2. From the experimental side, we provide additional experiments to compare the column-alone (uni-)clustering variant of the proposed approach with other methods, using synthetic data. We also provide additional experiments for phenotypic trait prediction and GWAS using image features rather than phenotypic traits as responses.

\section{Solution path for formulation 2}

Similar to the solution path we showed for formulation 1 in Section \ref{sec:solution_path}, we can obtain the solution path for formulation 2 as functions of two variables.

We first fix a reasonable $\lambda_1$ and vary $\lambda_2, \lambda_3$ to get solution paths for all the coefficients. These paths are shown in Figure \ref{path_f2_vary_lambda23}. The solution paths are smooth in $\lambda_2$ and $\lambda_3$. Similarly, we fix a reasonable $\lambda_2$ and vary $\lambda_1, \lambda_3$ to get solution paths for all the coefficients. These paths are shown in Figure \ref{path_f2_vary_lambda13}. The solution paths are smooth in $\lambda_1$ and $\lambda_3$. The \textit{reasonable} values are obtained using cross-validation.



\begin{figure*}[ht!]
\begin{center}
\begin{minipage}[t]{0.45\linewidth}
\centering
\includegraphics[width=0.95\textwidth]{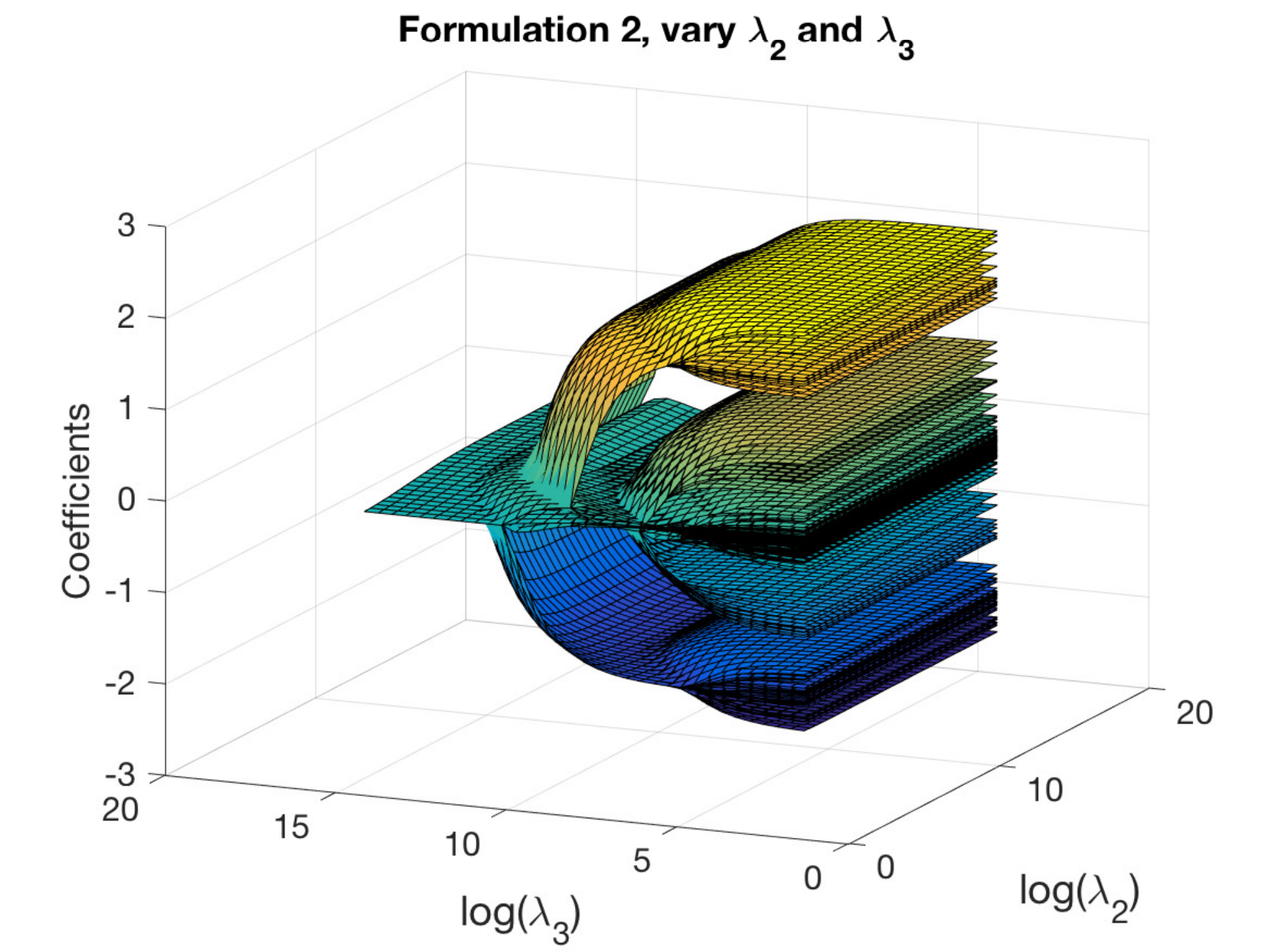}
\caption{Solution paths for formulation 2, fixing $\lambda_1$ and varying $\lambda_2, \lambda_3$.}
\label{path_f2_vary_lambda23}
\end{minipage}
\begin{minipage}[t]{0.45\linewidth}
\centering
\includegraphics[width=0.95\textwidth]{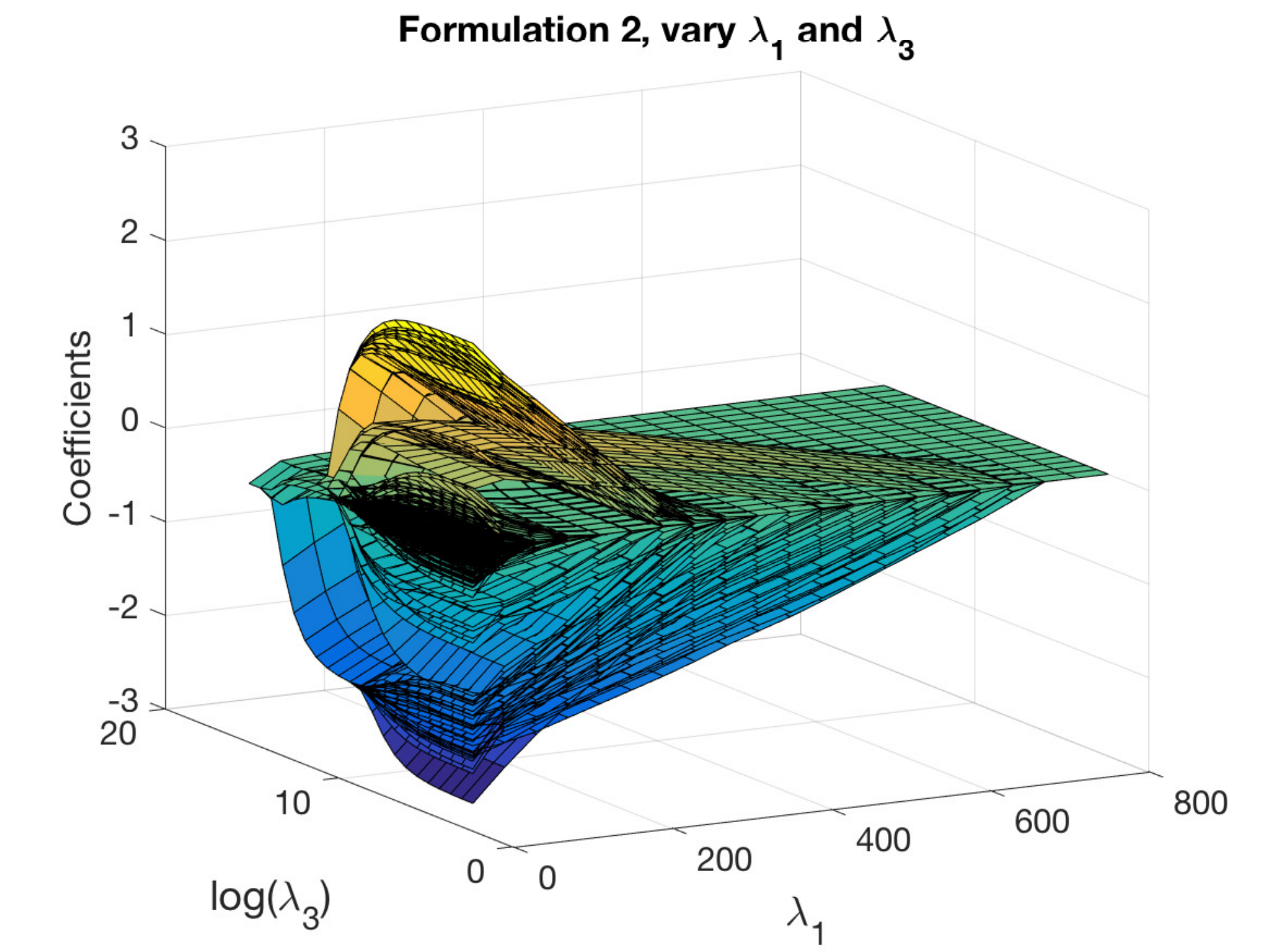}
\caption{Solution paths for formulation 2, fixing $\lambda_2$ and varying $\lambda_1, \lambda_3$.}
\label{path_f2_vary_lambda13}
\end{minipage}
\end{center}
\end{figure*}

\section{Definition of measures}

We provide definitions for the measures of quality of clustering used in synthetic data experiments (Section \ref{sec:simulation}).

\paragraph{F-1 score (F-1).} Assume $\mathcal B$ is the true clustering, define $TP$ to be the number of pairs of elements in $S$ that are in the same subset in $\mathcal A$ and in the same subset in $\mathcal B$. This is the true positive and similarly we can define $TN$, $FN$, $FP$ as true negative, false negative, and false positive, respectively. Define $precision = \frac{TP}{TP + FP}$ and $recall = \frac{TP}{TP + FN}$, the F-1 score is defined as:
\begin{equation}
\text{F-1} = \frac{2\cdot precision \cdot recall}{precision + recall}
\label{eqn:F1}
\end{equation}

\paragraph{Jaccard Index (JI).} Using the same notation as F-1 score, the Jaccard Index is defined as:
\begin{equation}
\text{JI} = \frac{TP}{TP + FP + FN}
\label{eqn:JI}
\end{equation}

\section{Optimization of the two formulations}

We provide the detailed update rules for optimizing formulation 1 described in Section \ref{subsec:opt_f1} here.

\subsection{Formulation 1 for multi-task regression}

\begin{itemize}
\item  Update for $f_1 = \| Y - X \Theta\|_F^2$: 
Let $(a_1; ...; a_k) = \text{prox}_{\sigma f_1}(b_1; ...; b_k),$
For each $s \in \{1,\ldots,k\}$, we have $$a_s = (\sigma X_s^TX_s + \frac 12 I_p)^{-1} \cdot (\sigma X_s^Ty_s + \frac 12 b_s).$$
This step corresponds to the closed-form formula of a ridge regression problem. For very large $p$ we can employ efficient approaches such as \cite{mcwilliams2014loco} and \cite{lu2014fast}.

\item Update for $f_2 = {\lambda_1  \sum_{i=1}^k \| \Theta_i\|_1}$: 
Let $(a_1; ...; a_k) = \text{prox}_{\sigma f_2}(b_1; ...; b_k),$
For each $s \in \{1,\ldots,k\}$, $j\in \{1,\ldots,p\}$,
$$[a_{s}]_j = \Bigg[ 1-\frac{\lambda_1\sigma}{ | [b_{s}]_j|} \Bigg]_{+} \cdot [b_{s}]_j.$$

\item Updates for $f_3 = \lambda_2 \Big[ \Omega_W(\Theta) + \Omega_{\tilde W}(\Theta^T) \Big]$: This is the standard bi-clustering problem on $\Theta$ and can be solved efficiently using the COnvex BiclusteRing  Algorithm (COBRA) introduced in \cite{chi2014convex}, and described in Algorithm \ref{alg:COBRA} for completeness.
 
\begin{algorithm}
\SetAlgoLined
\DontPrintSemicolon
\KwResult{Estimated $\Gamma$}
 Initialize $\Gamma_0 = \Theta$, $P_0 = 0$, $Q_0 = 0$, iteration $m = 0$ \;
 \While{not converged}{
   $Y_m = \textrm{prox}_{\frac{\lambda_3}{\lambda_2} \Omega_{\tilde W}} (\Gamma_m^T + P_m^T)$ (row clustering)\;
   $P_{m+1} = \Gamma_m + P_m -Y_m^T$\;
   $\Gamma_{m+1} = \textrm{prox}_{\frac{\lambda_3}{\lambda_2} \Omega_{W}} (Y_m^T + Q_m^T)$ (col. clustering)\;
   $Q_{m+1} = Y_m + Q_m -\Gamma_{m+1}^T$\;
   $m = m+1$\;
 }
 \caption{Convex biclustering (COBRA) \cite{chi2014convex}}
 \label{alg:COBRA}
\end{algorithm}

\end{itemize}

\subsection{Proof of Proposition \ref{prop:f1}}
\begin{proof}
We need to check that the conditions in Theorem 3.4 in \cite{combettes2008proximal} are satisfied in our case:
\begin{flalign*}
(\text{i}) &  \lim_{\|\Theta\| \to +\infty} f_1(\Theta)+ f_2(\Theta) + f_3(\Theta) = +\infty  &&\\
(\text{ii}) &  (0,\ldots,0) \in \text{sri}\{(\Theta-\Theta_{1}, \Theta-\Theta_{2}, \Theta-\Theta_{3}) | \Theta \in \mathbb R^{pk},\\
    & \qquad \Theta_{1} \in \text{dom}f_1,  \Theta_{2} \in \text{dom}f_2,\Theta_{3} \in \text{dom}f_{3}\} &&
\end{flalign*}
Let  $\mathcal H$ be the domain of $\Theta$ which can be set as $\mathbb R^{pk}$. Let $C$ be a nonempty convex subset of $\mathcal H$, the strong relative interior of $C$ is 
$$
\text{sri}(C) =  \big\{\Theta \in C | \text{cone}(C-\Theta) = \overline{\text{span}}(C-\Theta)\big\}
$$
where $\text{cone}(C) = \bigcup_{\lambda > 0}\{\lambda \Theta | \Theta \in C\}$, and $\overline{\text{span}}(C)$ is the closure of span $C$.

Now we check the conditions. For (i), $\|\Theta\|$ goes to infinity means some $\|\Theta_s\|$ goes to infinity, and then we know $f_2$ goes to infinity. Therefore (i) holds.

For (ii), we do not have any restriction on $\Theta$, so the right hand side is just $\text{sri}(\mathbb R^{pk})$, hence (ii) holds.


Therefore, the proposition follows according to Theorem 3.4 of \cite{combettes2008proximal}.

\end{proof}

\subsection{Proof of Proposition \ref{prop:f2}}
\begin{proof}
The optimization step for $\Gamma$ is solved by COBRA and it converges to global minimizer according to Proposition 4.1 in \cite{chi2014convex}. Define $f(\Theta, \Gamma) = \lambda_2 \sum_{i=1}^k \| \Theta_i - \Gamma_i\|_2^2$ with $\nabla f_\Theta(\Theta, \Gamma) = 2\lambda_2(\Theta - \Gamma)$ and $\nabla f_\Gamma(\Theta, \Gamma) = 2\lambda_2(\Gamma - \Theta)$, it is clear that $\nabla f_\Theta(\Theta, \Gamma)$ and $\nabla f_\Gamma(\Theta, \Gamma)$ are both Lipschitz-continuous in $\Theta$ and $\Gamma$, respectively. Since the optimization step for $\Theta$ is also assumed to find global minimizer, Theorem 3.9 in \cite{beck2015convergence} guarantees that our algorithm in Section \ref{subsec:opt_f2} converges to the global minimizer.
\end{proof}

\section{Synthetic data experiments for \emph{uni-clustering}}

\begin{figure*}
\minipage{0.4\textwidth}
\begin{center}
\centerline{\includegraphics[width=\columnwidth]{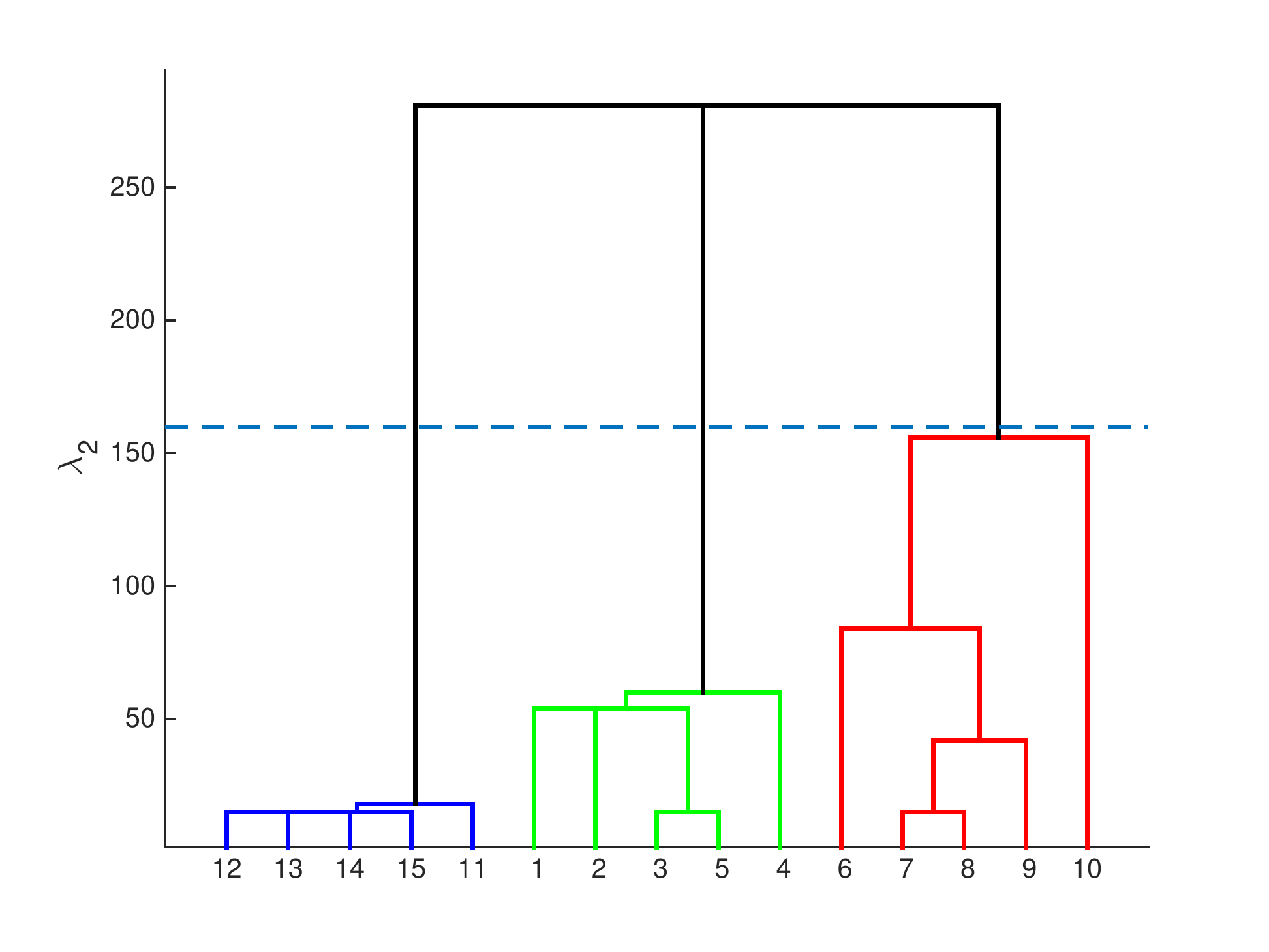}}
\caption{Tree structure learnt by our method}
\label{fig:example}
\end{center}
\endminipage\hfill
\minipage{0.4\textwidth}
\begin{center}
\centerline{\includegraphics[width=\columnwidth]{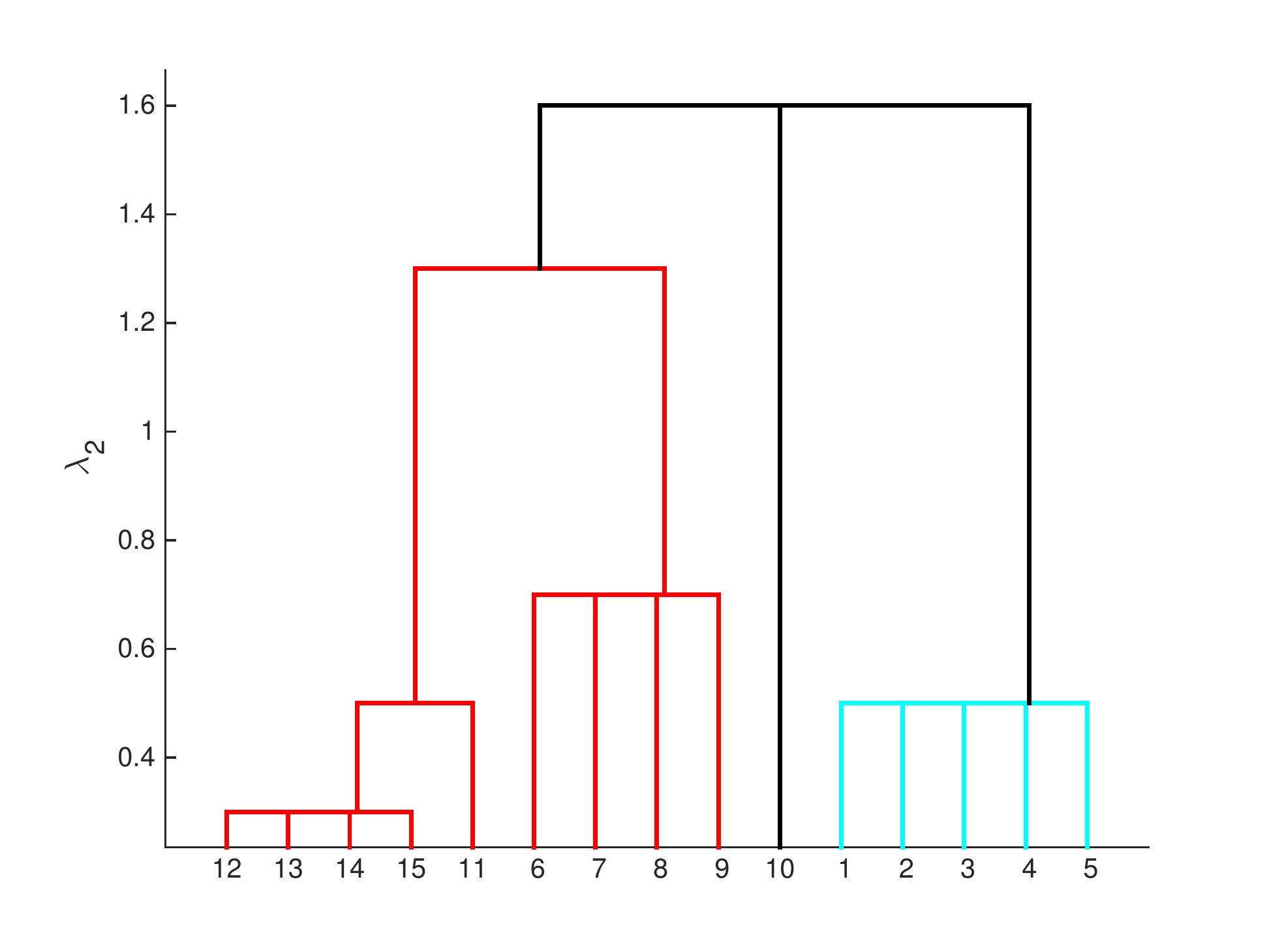}}
\caption{Tree structure learnt by post-clustering of Single Task Lasso}
\label{fig:examplesingle}
\end{center}
\endminipage\hfill

\end{figure*}

In this section we run additional experiments on synthetic data for column-alone clustering (\textit{uni-clustering}) variant of formulation 1. To display the hierarchical tree structure, we use relatively small scale data here. We compare our methods with the following approaches.
\begin{itemize}
\item {\bf Single task}: Each task is learned separately via Lasso.
\item {\bf No group MTL}~\cite{obozinski2006multi}: The traditional multitask approach using group lasso penalty, where all tasks are learned jointly and the same features are selected across tasks.
\item {\bf Pre-group MTL}: Given the \emph{true} number of groups, first partition the tasks purely according to the correlation among responses and then apply \emph{No group MTL} in each cluster.
\item {\bf Kang \textit{et al.}}~\cite{kang2011learning}: Mixed integer program learning a shared feature representations among tasks, while simultaneously determining ``with whom'' each task should share. We used the code provided by the authors of~\cite{kang2011learning} and used the true number of tasks.
\item {\bf Tree-guided group Lasso}~\cite{kim2010tree}: Employs a structured penalty function induced from a predefined tree structure among responses, that encourages multiple correlated responses to share a similar set of covariates. We used the code provided by the authors of~\cite{kim2010tree} where the tree structure is obtained by running a hierarchical agglomerative clustering on the responses.
\end{itemize}

We consider $n=20$ samples, $p = 50$ features, and three groups of tasks. Within each group there are 5 tasks whose parameter vectors are sparse and identical to each other. We generate independent train, validation, and test sets. For each method, we select the regularization parameters using the validation sets, and report the root-mean-squared-error (RMSE)  of the resulting models on the test sets. We repeat this procedure 5 times. The results are reported in Table \ref{RMSE}.  From the table we can see that the \emph{Single task} method has the largest RMSE as it does not leverage task relatedness; \emph{No group MTL} has slightly better RMSE; both \emph{Kang et al.} and \emph{Tree guided group Lasso} get some improvement by considering structures among tasks; \emph{Pre-group MLT} achieves a better result, mainly because it is given the true number of groups and for these synthetic datasets it is quite easy to obtain good groups via response similarity, which might not necessarily be the case with real data and when the predictors differ from task to task. Our method achieves the smallest RMSE, outperforming all  approaches compared.
We also report the running times of each method, where we fix the error tolerance to $10^{-5}$ for fair comparisons. Though the timing for each method could always be improved using more effective implementations, the timing result reflect the algorithmic simplicity of the steps in our approach compare to e.g. the mixed integer program of~\cite{kang2011learning}. 

\begin{table}[t]
\caption{RMSE for different comparison methods}
\label{RMSE}
\vskip 0.15in
\begin{center}
\begin{small}
\begin{tabular}{lccc}
\hline
Method & RMSE & std & time\\
\hline
Single task &  5.665 & 0.131 & 0.02\\
No group multitask learning &  5.520 & 0.115 & 0.05\\
Pre-group multitask learning &  5.256 & 0.117 & 0.10\\
Kang et al & 5.443 & 0.096 & $>10$\\
Tree guided group Lasso &  5.448 & 0.127 & 0.03\\
Ours &  \bf{4.828} & 0.117 & 0.16\\
\hline
\end{tabular}
\end{small}
\end{center}
\vskip -0.1in
\end{table}

An example of tree structure learnt by our approach is shown in Figure~\ref{fig:example}. It is contrasted with Figure~\ref{fig:examplesingle}, which depicts the structure obtained \emph{a-posteriori} by performing hierarchical clustering on the regression matrix learnt by \emph{Single Task.} The simulation scenario is the same as before except that the non-zero coefficients in each true group are not equal but sampled as $0.5 + N(0,1)/3,$ where $N$ denotes the normal distribution.  As can be seen from Figure~\ref{fig:example}, no matter what $\lambda_2$ is, our approach never makes a mistake in the sense that it never puts tasks from different true groups in the same cluster. For $\lambda_2>150$ our approach recognizes the true groups. As $\lambda_2$ becomes very large there are no intermediary situations where two tasks are merged first. Instead all tasks are put in the same cluster.  We see tasks $\{3,5\}$ merge first in group $\{1-5\}$ and $\{7,8,9\}$ merge first in group $\{6-10\}.$ This corresponds to the fact that tasks $\{3,5\}$ have largest correlation among group $\{1-5\}$ and $\{7,8,9\}$ has largest correlation among group $\{6-10\}.$
We can see in Figure~\ref{fig:examplesingle} that for \emph{Single Task} post clustering, task 10 does not merge with $\{6-9\}.$


%

\paragraph{Impact of the weights $w_{ij}.$}
One might argue that our approach relies on ``good'' weights $w_{ij}$ among tasks. However, it turns out that it is fairly robust to the weights. 
Recall that we select the weight $w_{ij}$ by $\kappa$-nearest-neighbors. In this synthetic dataset, we have 5 tasks in each group so the most natural way is to set $\kappa = 4$. We also try setting $\kappa = 2,3,5,6$ and see how this affects the result. The test RMSEs for different $\kappa$'s are given in Table \ref{kappa}.
From the table we see that although the best performance is when we select $\kappa = 4$, our method is quite robust to the choice of weights, especially when $\kappa$ is smaller than the natural one. 
When $\kappa$ is large the result gets slightly worse, because now we cannot avoid positive weights across groups. But even in this case, our method is still competitive.

\begin{table}[!h]
\small
\caption{RMSE for our approach with weight specification by $\kappa$- nearest-neighbors.}
\label{kappa}
\vskip -0.15in
\begin{center}
\begin{small}
\begin{tabular}{lccccc}
\hline
$\kappa$ & 2 & 3 & 4 & 5 & 6 \\
\hline
RMSE &  4.847 & 4.836 & \bf{4.828} & 4.896 &4.928 \\
\hline
\end{tabular}
\end{small}
\end{center}
\vskip -0.1in
\end{table}

\section{Real Data Analysis on uni-clustering structure}
We provide additional real data experiments with the column-alone (uni-)clustering variant of our approach to augment those in Section \ref{sec:real_data}. The varieties and their names used in the trait prediction experiments (Section \ref{trait_prediction_expt} and Section  \ref{trait_prediction_expt_2}) are given in Table \ref{tab:variety_names}. We provide an experiment that directly tries to associate remotely sensed image features with SNP data. In this case, since the image features are likely multi-dimensional, our approach of learning the group structure among the feature dimensions is intuitive. Note that this setting is different from GWAS, where we associate the traits with SNP data. 

\begin{table}
\begin{tabular}{ |c|l| } 
 \hline
 Variety number & Variety name \\ 
 \hline
 1 & RS 392x105 BMR FS \\ 
 2 & RS 400x38 BMR SG \\ 
 3 & RS 341x10 FG white \\ 
 4 & RS 374x66 FS \\ 
 5 & RS 327x36 BMR FS \\ 
 6 & RS 400x82 BMR SG \\ 
 7 & RS 366x58 FG white \\ 
 8 & SP NK5418 GS \\ 
 9 & SP NK8416 GS \\ 
 10 & SP SS405 FS \\ 
 11 & SP Trudan Headless FS PS \\ 
 12 & SP Trudan 8 FS \\ 
 13 & SP HIKANE II FS \\ 
 14 & SP NK300 FS \\ 
  15 & SP Sordan Headless FS PS \\ 
 16 & SP Sordan 79 FS \\ 
 17 & PH 849F FS \\ 
 18 & PH 877F FS \\ 
 \hline
\end{tabular}
\caption{Numbers and names of Sorghum varieties used in experiments.}
\label{tab:variety_names}
\end{table}

\subsection{Phenotypic Trait Prediction from Remote Sensed Data}
\label{trait_prediction_expt_2}
We repeat the experiment described in Section \ref{trait_prediction_expt} for stalk diameter and stalk volume traits and provide the tree structure for tasks in Figures \ref{Ph2} and \ref{Ph3}. They look very similar to each other and this makes sense since stalk diameter and stalk volume are highly correlated. For these two traits, we can see that variety 12 is very different from others. It corresponds to tall thin plants with few small dark leaves.
 

%
%

\subsection{GWAS dataset}
\label{GWAS_expt}
In this experiment, we use SNP data from 850 varieties as input ($X$).  We considered $p=3025$ SNPs (features). There are $n=1920$ plots (observations), each containing a single variety.  The output data ($Y$) is the histogram of photogrammetrically derived heights obtained from RGB images of the $n=1920$ plots. We consider $10$ bins, and each bin is treated as a task. Therefore,  $k=10$. It has been demonstrated that height histograms describe the structure of the plants in the plot and are hence powerful predictors of various traits \cite{ramamurthy2016predictive}. Therefore it is worthwhile performing genomic mapping using them bypassing trait prediction. Since it is reasonable to expect the neighboring bins of the histograms to be correlated, our approach for hierarchical task grouping will result in an improved association discovery. 

For this dataset, Kang \textit{et al.}'s method \cite{kang2011learning} did not scale to handle the amount of features. In general, we noticed that this algorithm is quite unstable, namely the task membership kept changing at each iteration especially as the dimensionality increases. 

The tree structure obtained by our method is given in Figure \ref{GWAS}. Please note that the $y$-axis in the figure is $\log(\lambda_2).$ We notice that bins $\{8,9,10\}$ merge quickly while bins \{2,3,4\} merge when $\lambda_2$ is extremely large. 
Note that the distance from bin 5 to bin 4 is much larger, compared to the distance from bin 7 to bin 5. In the figure these look similar due to logarithmic scale. 
Bins $\{1,7,8,9,10\}$ are rarely populated. They all have small coefficients and merge together quickly, while more populated bins tend to merge later. 

\begin{figure}[t]`
\begin{center}
\centerline{\includegraphics[width=\columnwidth]{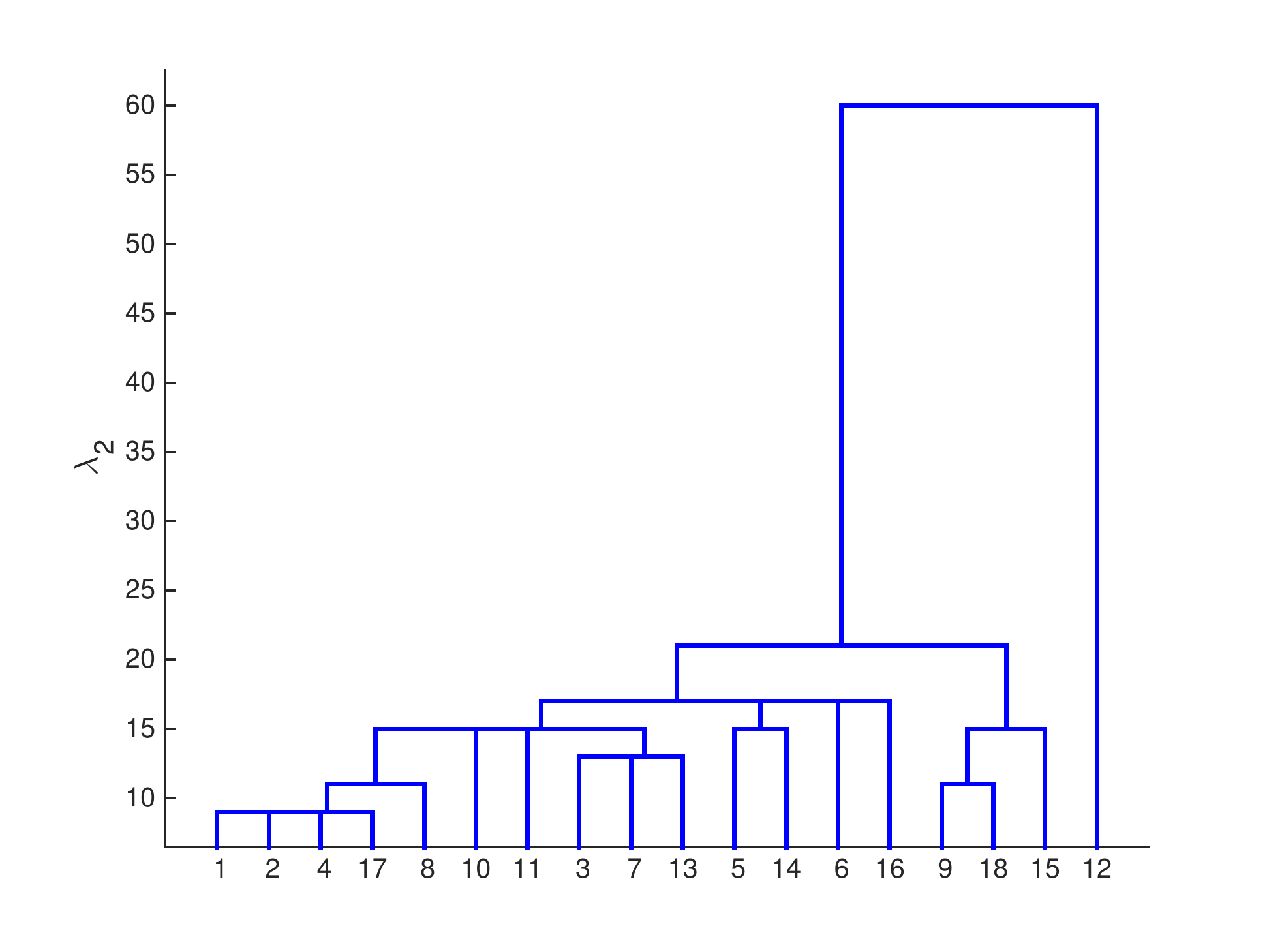}}
\caption{Tree structure of tasks (varieties) inferred using our approach for stalk diameter.}
\label{Ph2}
\end{center}
\end{figure}

\begin{figure}[t]
\begin{center}
\centerline{\includegraphics[width=\columnwidth]{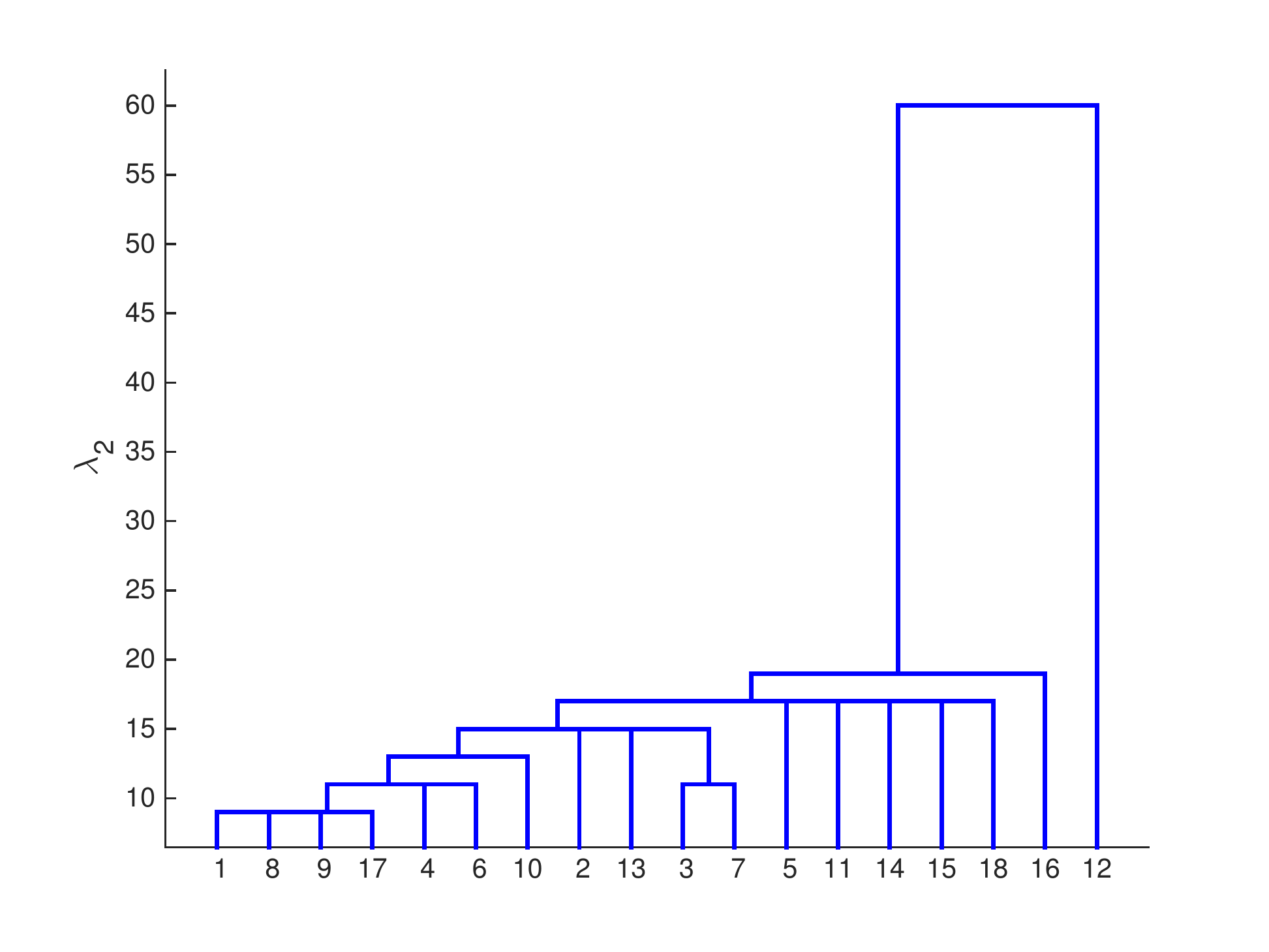}}
\caption{Tree structure of tasks (varieties) inferred using our approach for stalk volume.}
\label{Ph3}
\end{center}
\end{figure}

\begin{figure}[t]
\begin{center}
\centerline{\includegraphics[width=\columnwidth]{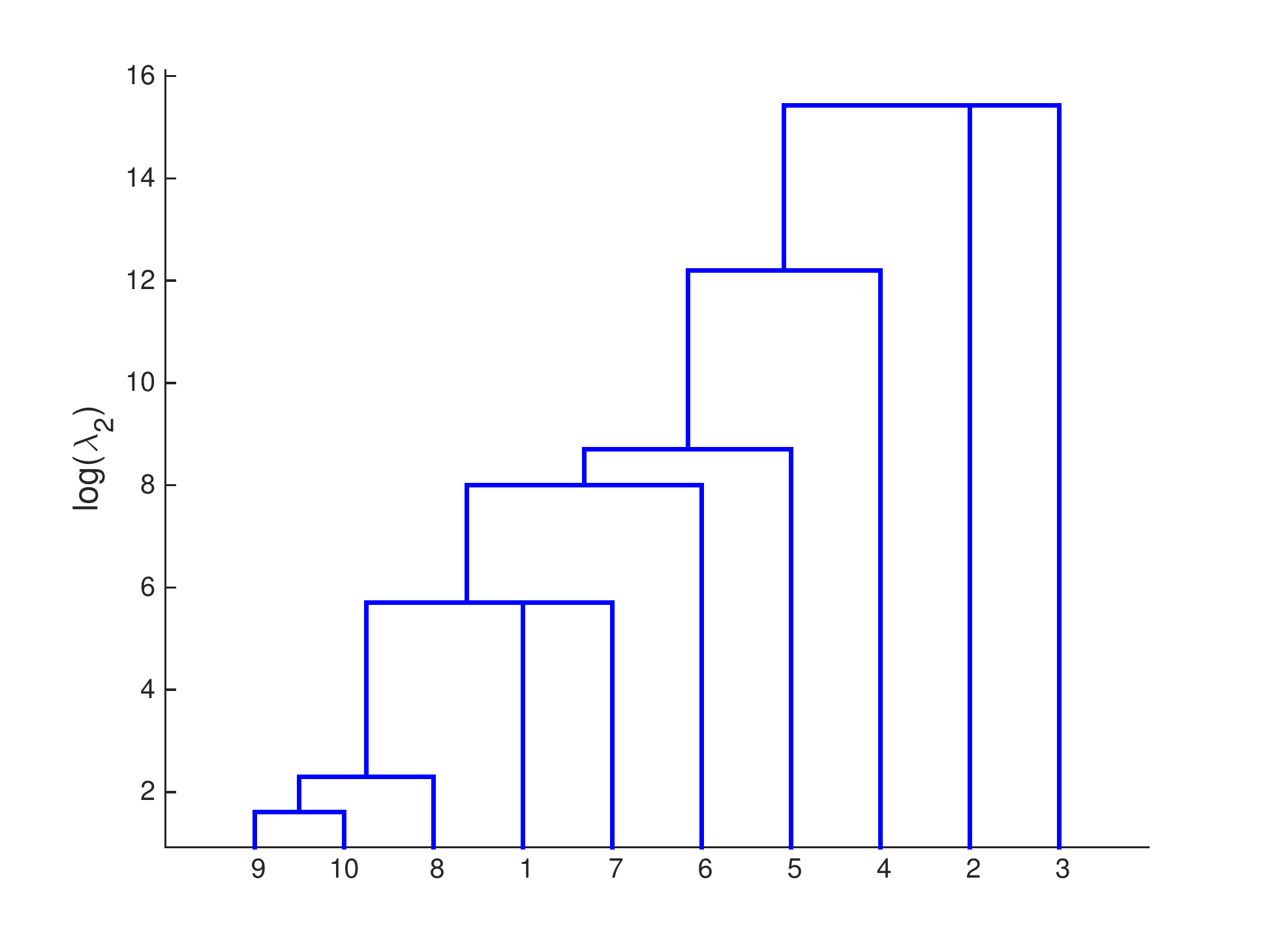}}
\caption{Tree structure of the tasks (height bins) inferred using our approach for the GWAS dataset.}
\label{GWAS}
\end{center}
\end{figure}

\end{document}